\documentclass{article}

\PassOptionsToPackage{round}{natbib}

 \usepackage[preprint]{neurips_2026}

\usepackage[utf8]{inputenc} %
\usepackage[T1]{fontenc}    %
\usepackage{hyperref}       %
\usepackage{url}            %
\usepackage{booktabs}       %
\usepackage{amsfonts}       %
\usepackage{nicefrac}       %
\usepackage{microtype}      %
\usepackage{xcolor}         %
\usepackage{siunitx}
\usepackage{soul}
\usepackage{subcaption}
\usepackage{fontawesome5}
\usepackage{graphicx}
\usepackage{float}
\usepackage{algorithm}
\usepackage{algpseudocode}

\usepackage{pifont}%
\usepackage{amsmath,amsthm,amssymb,mathtools}
\usepackage{makecell}
\usepackage{wrapfig}
\usepackage[dvipsnames]{xcolor}

\title{Modular Multimodal Classification Without Fine-Tuning: A Simple Compositional Approach}

\author{%
  Herman Bergström\thanks{Equal contribution. $^\dag$Work was performed while at the University of Toronto.}$^{\,\,\,\dag 1, 2}$ \quad
  Aditya Mehrotra$^{* 2, 3}$ \quad
  Rahul G. Krishnan$^{2, 3}$ \\
  \\
  $^1$Chalmers University of Technology and University of Gothenburg \\
  $^2$Vector Institute \quad
  $^3$University of Toronto \\
  \texttt{hermanb@chalmers.se} \quad \texttt{aditmeh@cs.toronto.edu}
}

\newcommand{\hltodo}[1]{}
\newcommand{\hladitya}[1]{}

\begin{document}

\def\alt{\text{Alt}}
\def\kl{\text{KL}}

\newtheorem{thmlem}{Lemma}
\newtheorem{thmcol}{Corollary}
\newtheorem{thmconj}{Conjecture}
\newtheorem{thmprop}{Proposition}
\newtheorem{thmthm}{Theorem}
\newtheorem{thmappthm}{Theorem}
\newtheorem{thmasmp}{Assumption}
\newtheorem{thmcorr}{Corollary}
\theoremstyle{definition}
\newtheorem{thmdef}{Definition}
\newtheorem{thmex}{Example}
\newtheorem{thmrem}{Remark}
\newtheorem*{thmrem*}{Remark}

\newenvironment{proofsketch}{%
  \renewcommand{\proofname}{Proof sketch}\proof}{\endproof}

\newcommand\independent{\protect\mathpalette{\protect\independenT}{\perp}}
\def\independenT#1#2{\mathrel{\rlap{$#1#2$}\mkern2mu{#1#2}}}
\def\indep{\independent{}}

\def\R{\mathbb{R}}
\def\Z{\mathbb{Z}}
\def\E{\mathbb{E}}
\def\V{\mathbb{V}}
\def\hQ{\hat{Q}}
\def\hW{\hat{W}}
\def\hw{\hat{w}}
\def\hy{\hat{y}}
\def\hf{\hat{f}}
\def\hg{\hat{g}}
\def\hh{\hat{h}}
\def\htheta{\hat{\theta}}

\def\ba{{\bf a}}
\def\bb{{\bf b}}
\def\bc{{\bf c}}
\def\bd{{\bf d}}
\def\be{{\bf e}}
\def\bg{{\bf g}}
\def\bh{{\bf h}}
\def\bi{{\bf i}}
\def\bj{{\bf j}}
\def\bk{{\bf k}}
\def\bl{{\bf l}}
\def\bm{{\bf m}}
\def\bn{{\bf n}}
\def\bo{{\bf o}}
\def\bp{{\bf p}}
\def\bq{{\bf q}}
\def\br{{\bf r}}
\def\bs{{\bf s}}
\def\bt{{\bf t}}
\def\bu{{\bf u}}
\def\bv{{\bf v}}
\def\bw{{\bf w}}
\def\bx{{\bf x}}
\def\by{{\bf y}}
\def\bz{{\bf z}}
\def\bA{{\bf A}}
\def\bB{{\bf B}}
\def\bC{{\bf C}}
\def\bD{{\bf D}}
\def\bE{{\bf E}}
\def\bF{{\bf F}}
\def\bG{{\bf G}}
\def\bH{{\bf H}}
\def\bI{{\bf I}}
\def\bJ{{\bf J}}
\def\bK{{\bf K}}
\def\bL{{\bf L}}
\def\bM{{\bf M}}
\def\bN{{\bf N}}
\def\bO{{\bf O}}
\def\bP{{\bf P}}
\def\bQ{{\bf Q}}
\def\bR{{\bf R}}
\def\bS{{\bf S}}
\def\bT{{\bf T}}
\def\bU{{\bf U}}
\def\bV{{\bf V}}
\def\bW{{\bf W}}
\def\bX{{\bf X}}
\def\bY{{\bf Y}}
\def\bZ{{\bf Z}}
\def\bba{\mathbb{a}}
\def\bbb{\mathbb{b}}
\def\bbc{\mathbb{c}}
\def\bbd{\mathbb{d}}
\def\bbe{\mathbb{e}}
\def\bbf{\mathbb{f}}
\def\bbg{\mathbb{g}}
\def\bbh{\mathbb{h}}
\def\bbi{\mathbb{i}}
\def\bbj{\mathbb{j}}
\def\bbk{\mathbb{k}}
\def\bbl{\mathbb{l}}
\def\bbm{\mathbb{m}}
\def\bbn{\mathbb{n}}
\def\bbo{\mathbb{o}}
\def\bbp{\mathbb{p}}
\def\bbq{\mathbb{q}}
\def\bbr{\mathbb{r}}
\def\bbs{\mathbb{s}}
\def\bbt{\mathbb{t}}
\def\bbu{\mathbb{u}}
\def\bbv{\mathbb{v}}
\def\bbw{\mathbb{w}}
\def\bbx{\mathbb{x}}
\def\bby{\mathbb{y}}
\def\bbz{\mathbb{z}}
\def\bbA{\mathbb{A}}
\def\bbB{\mathbb{B}}
\def\bbC{\mathbb{C}}
\def\bbD{\mathbb{D}}
\def\bbE{\mathbb{E}}
\def\bbF{\mathbb{F}}
\def\bbG{\mathbb{G}}
\def\bbH{\mathbb{H}}
\def\bbI{\mathbb{I}}
\def\bbJ{\mathbb{J}}
\def\bbK{\mathbb{K}}
\def\bbL{\mathbb{L}}
\def\bbM{\mathbb{M}}
\def\bbN{\mathbb{N}}
\def\bbO{\mathbb{O}}
\def\bbP{\mathbb{P}}
\def\bbQ{\mathbb{Q}}
\def\bbR{\mathbb{R}}
\def\bbS{\mathbb{S}}
\def\bbT{\mathbb{T}}
\def\bbU{\mathbb{U}}
\def\bbV{\mathbb{V}}
\def\bbW{\mathbb{W}}
\def\bbX{\mathbb{X}}
\def\bbY{\mathbb{Y}}
\def\bbZ{\mathbb{Z}}
\def\ca{\mathcal{a}}
\def\cb{\mathcal{b}}
\def\cc{\mathcal{c}}
\def\cd{\mathcal{d}}
\def\ce{\mathcal{e}}
\def\cf{\mathcal{f}}
\def\cg{\mathcal{g}}
\def\ch{\mathcal{h}}
\def\ci{\mathcal{i}}
\def\cj{\mathcal{j}}
\def\ck{\mathcal{k}}
\def\cl{\mathcal{l}}
\def\cm{\mathcal{m}}
\def\cn{\mathcal{n}}
\def\co{\mathcal{o}}
\def\cp{\mathcal{p}}
\def\cq{\mathcal{q}}
\def\cs{\mathcal{s}}
\def\ct{\mathcal{t}}
\def\cu{\mathcal{u}}
\def\cv{\mathcal{v}}
\def\cw{\mathcal{w}}
\def\cx{\mathcal{x}}
\def\cy{\mathcal{y}}
\def\cz{\mathcal{z}}
\def\cA{\mathcal{A}}
\def\cB{\mathcal{B}}
\def\cC{\mathcal{C}}
\def\cD{\mathcal{D}}
\def\cE{\mathcal{E}}
\def\cF{\mathcal{F}}
\def\cG{\mathcal{G}}
\def\cH{\mathcal{H}}
\def\cI{\mathcal{I}}
\def\cJ{\mathcal{J}}
\def\cK{\mathcal{K}}
\def\cL{\mathcal{L}}
\def\cM{\mathcal{M}}
\def\cN{\mathcal{N}}
\def\cO{\mathcal{O}}
\def\cP{\mathcal{P}}
\def\cQ{\mathcal{Q}}
\def\cR{\mathcal{R}}
\def\cS{\mathcal{S}}
\def\cT{\mathcal{T}}
\def\cU{\mathcal{U}}
\def\cV{\mathcal{V}}
\def\cW{\mathcal{W}}
\def\cX{\mathcal{X}}
\def\cY{\mathcal{Y}}
\def\cZ{\mathcal{Z}}

\maketitle

\begin{abstract}

We introduce CoMET, \textit{\textbf{C}omposing \textbf{M}odality \textbf{E}ncoders with \textbf{T}abular foundation models}, a simple yet highly competitive method for multimodal classification: pass each modality through a frozen pre-trained backbone, compress the resulting embeddings with PCA, and concatenate as input into a Tabular Foundation Model (TFM) for prediction. We show that PCA alone suffices to act as an adaptor yielding strong, robust performance across modalities. When the \texttt{CLS} tokens of the foundation model align poorly with downstream tasks, we propose \textbf{PALPooling}, a lightweight adaptive token pooler that consistently improves representation quality. By composing strong frozen representation learning backbones with TFMs, our approach achieves state-of-the-art results across diverse multimodal benchmarks without any training. On hierarchical tasks with large fine-grained class spaces, our approach enables fast and scalable classification, handling datasets with over 500,000 samples and 2,000 classes without any fine-tuning. Overall, our results show that the composition of foundation models is a simple, yet powerful, out-of-the-box solution for multimodal learning, challenging the necessity of complex, end-to-end training pipelines for new problems.

\end{abstract}

\section{Introduction}

Tabular Foundation Models (TFMs) have recently demonstrated that strong predictive performance can be achieved by amortizing Bayesian inference via in-context learning~\citep{qu2025tabicl, hollmann2022tabpfn, ma2024tabdpt}. After pre-training, these models can be applied to new datasets without additional fine-tuning, avoiding the substantial training times that come with task-specific pipelines.
Surprisingly, despite being trained on priors designed to mimic the data-generating process observed in tabular data, the models are effectively data-agnostic. Instead, we find that, by applying simple dimensionality reduction techniques, TFMs become competitive classification heads for representations extracted using modality-specific encoders, such as image and text (see section \ref{sec:PCA}). 

This unexpected modality agnosticism raises a natural question: Can TFMs be used for fine-tuning-free multimodal classification? While there is limited initial work leveraging TFMs in multimodal setups, these either employ dataset-specific fine-tuning~\citep{kim2026multimodalpfn}, or treat them as tabular encoders operating solely on tabular data~\citep{luo2025time}. We argue that this not only detracts from the out-of-the-box nature of these models, but it is also not necessary to achieve competitive performance. Instead, we propose solving the task as a simple composition of foundation models, leveraging pre-trained encoders~\citep{simeoni2025dinov3, clark2020electra} for non-tabular features, and a TFM to make the final predictions, a modular pipeline where components can be independently replaced or extended without retraining.

Despite this simplicity, a potential bottleneck emerges in the interface between modality encoders and TFMs. In practice, representations are typically obtained via fixed pooling strategies, such as \texttt{CLS}-tokens or mean-pooling~\citep{dosovitskiy2020image, devlin2019bert}. When these embeddings fail to prioritize features relevant to the classification task, the final prediction quality may be limited, even when the backbone and TFM are individually strong. While adaptive pooling of individual tokens can traditionally help alleviate this issue~\citep{reimers2019sentence}, this would require fine-tuning, which in our setting would entail backpropagation through the TFM. To address this, we introduce Pseudo Attention Label Pooling (PALPooling). PALPooling is a lightweight pooling method that can be fit in seconds using one to three forward passes, preserving the training-free pipeline while enabling adaptive pooling in settings where backpropagation is not possible.

In this work, we make the following contributions. \textbf{i)} We demonstrate that reducing the embedding dimensionality through PCA is enough to make TabICL a competitive multimodal classifier, without any further fine-tuning. \textbf{ii)} Given this finding, we propose \textbf{CoMET}, \textit{Composing Modality Encoders with Tabular foundation models}, an easy-to-implement framework offering high classification accuracy with minimal implementation effort. \textbf{iii)} We introduce PALPooling, an adaptive pooling method that uses pseudo-attention labels to avoid backpropagation. The method demonstrates benefits for both image and text settings, outperforming \texttt{CLS} tokens that are misaligned with the target prediction, with minimal computational overhead. \textbf{iv)} We highlight that strong fine-tuning-free classification enables new applications, showcasing this in the setting of hierarchical prediction.

\section{Related Work}
\paragraph{Tabular Foundation Models} Prior-fitted networks were introduced by \citet{muller2021transformers}, and have since paved the way for a new approach to generalizable tabular foundation models~\citep{hollmann2022tabpfn}. The typical approach, employed by models like  TabPFN~\citep{hollmann2025accurate} and TabICL~\citep{qu2025tabicl}, entails fitting the posterior predictive distribution on extensive synthetically generated data. However, models such as TabDPT~\citep{ma2024tabdpt} show that competitive performance can be achieved when using large amounts of real-world data for the prior. While these TFMs were originally limited to small datasets with few features, recent iterations have begun to relax these constraints~\citep{qu2026tabiclv2, grinsztajn2025tabpfn}. Although the focus of this paper is to showcase the proof of concept, we anticipate that the underlying framework will naturally extend to other tasks such as causal effect estimation \cite{balazadeh2025causalpfn}.

\paragraph{Multimodal Classification} While there is extensive work exploring architectures that can efficiently integrate tabular features with other modalities, these often focus on a single non-tabular modality at a time. \citet{bonnier2024revisiting} propose TTT, a Tabular Text Transformer which constructs embedding vectors through a distance-to-quantile mechanism, but requires dataset-specific fine-tuning. TabSTAR~\citep{arazi2025tabstar} is closer to a pre-trained foundation model, but still requires a small amount of dataset-specific fine-tuning using Low Rank Adaptations~\citep{hu2022lora}.
Another line of research utilizes contrastive learning between tabular and image features to learn joint representations, with models such as TIP~\citep{du2024tip} and MMCL~\citep{Hager_2023_CVPR}. Still, these models require end-to-end fine-tuning on a per-dataset basis.

\paragraph{Tabular Foundation Models on Multimodal Data} TIME~\citep{luo2025time} uses TabPFN to extract representations for the tabular features, and subsequently fuses these with image features (e.g., through concatenation) before training a linear classifier on top. Perhaps most similar to our setting, the recently proposed MMPFN~\citep{kim2026multimodalpfn}, which forwards image and text features through the TabPFN backbone, but crucially employs fine-tuning with modality-specific projectors and a tabular alignment loss. As we show in Section \ref{sec:results}, this is not required, and we outperform their model using simple dimensionality reduction and TabICLv2.

\section{Background}

Transformer-based foundation models have learned to encode rich structure across different data modalities~\citep{mur2026v, simeoni2025dinov3, clark2020electra}, outputting contextualized token embeddings. Concretely, given an input from modality $m$, the backbone $\Phi_m(\cdot)$ produces a set of token embeddings corresponding to modality-specific units (e.g., image patches or subword tokens). To aggregate these tokens into a fixed-size representation suitable for downstream tasks, we typically leverage a pooling function, such as mean-pooling, learned attention-based pooling, or direct use of a dedicated classification (\texttt{CLS})-token, a special token trained to summarize the input according to the model’s pre-training objective.

We focus on the task of multiclass classification. Formally, for a sample $i$, we wish to predict the class label $y_i \in \{1, ..., C \}$ based on features $\mathbf{x}_i \in \R^d$. $\mathbf{x}_i$ can correspond to features from a single modality, or a concatenation of multiple (e.g., $\mathbf{x}_i = (\mathbf{x}^{tab}_i, \mathbf{x}^{img}_i, \mathbf{x}^{text}_i)$). For any non-tabular modality, we leverage the encoders mentioned above to extract $P$ local token embeddings $\mathbf{X}_i^{m} \in \R^{P \times d_m}$, and apply a pooling function, $\mathrm{Pool}(\mathbf{X}_i^{m})$, to construct summarized embeddings $\mathbf{x}^{m}_i \in \R^{d_m}$.

TFMs are amortized meta-learners~\citep{qu2025tabicl, liutabpfn}, trained to predict the posterior predictive distribution over a set of \emph{query} features, conditioned on a \emph{support} set, $\mathcal{D}_{\text{sup}} = \{(\mathbf{x}_1, y_1), \dots, (\mathbf{x}_n, y_n) \}$. By training solely on synthetically generated data, they approximate $ \hat{\mathbf{y}}_{qry} \approx P(y_{qry} | \mathbf{x}_{qry}, \mathcal{D}_{\text{sup}})$, where $\hat{\mathbf{y}}_{qry} \in [0,1]^C$ corresponds to the predicted label distribution. While $\mathcal{D}_{\text{sup}}$ and $\mathcal{D}_{\text{qry}}$ will in many cases correspond to the training and evaluation sets, respectively, TFMs leverage the support set through in-context learning, meaning no parameters are tuned. Note that since we distinguish between token-level representations $\mathbf{X}_i^{m}$ and pooled features $\mathbf{x}^{m}_i$, we let $\mathcal{D}_{\text{sup}}$ correspond to pooled multimodal features by default, and use a superscript $\mathcal{D}^{m, \text{tok}}_{\text{sup}} = \{\mathbf{X}^m_i, y_i  \}_{i \in \mathcal{I}_{\text{sup}}}$, when operating on modality-specific token level representations in Section \ref{sec:PALPooling}.

\section{Method}

When making a multimodal prediction, the \textbf{CoMET} method works as follows: \textbf{1.} For each non-tabular modality, we extract sample-wise token-level embeddings $\mathbf{X}^m_i$ using a frozen pre-trained backbone $\Phi_m$. \textbf{2.} Vector representations $\mathbf{x}^{m}_i$ are constructed using a (potentially learned) pooling function $\mathrm{Pool}(\mathbf{X}^m_i)$. \textbf{3.} We fit a PCA projection $\mathbf{U}_m \in \R^{d_m \times \tilde{d}_m}$ on the pooled features in $\mathcal{D}_{sup}$ (i.e., on $\{ \mathbf{x}^m_i \}_{i \in \mathcal{I}_{\text{sup}}}$), giving us the compressed modality-specific representations  $\tilde{\mathbf{x}}^m_i = \mathbf{U}_m^\top\mathbf{x}^m_i$. \textbf{4.} The projected support set $\tilde{\mathcal{D}}_{sup} = \{ (\tilde{\mathbf{x}}_i, y_i) \}_{i \in \mathcal{I}_{\text{sup}}}$, where $\tilde{\mathbf{x}}_i = (\mathbf{x}^{tab}_i, \tilde{\mathbf{x}}^{m_1}_i, \dots , \tilde{\mathbf{x}}^{m_M}_i)$, is then passed to the TFM to make the final prediction, $\hat{\mathbf{y}}_{qry} = \mathrm{TFM}(\tilde{\mathbf{x}}_{qry}, \tilde{\mathcal{D}}_{sup})$

\subsection{PALPooling}
\label{sec:PALPooling}

Adaptive pooling of image and text tokens often outperforms relying on the \texttt{CLS}-token~\citep{reimers2019sentence}. Still, in our setting, traditional fine-tuning of such a layer would require backpropagation through a tabular foundation model, detracting from its off-the-shelf nature. To avoid this, we introduce Pseudo Attention Label Pooling (PALPooling). Inspired by the fact that a standard attention layer with one head and a learned query vector simplifies to linear scoring, PALPooling fits this linear model through ridge regression on token-level pseudo attention labels derived from the model output, as opposed to backpropagation.

\begin{figure}[t]
    \centering
    \includegraphics[width=0.98\linewidth]{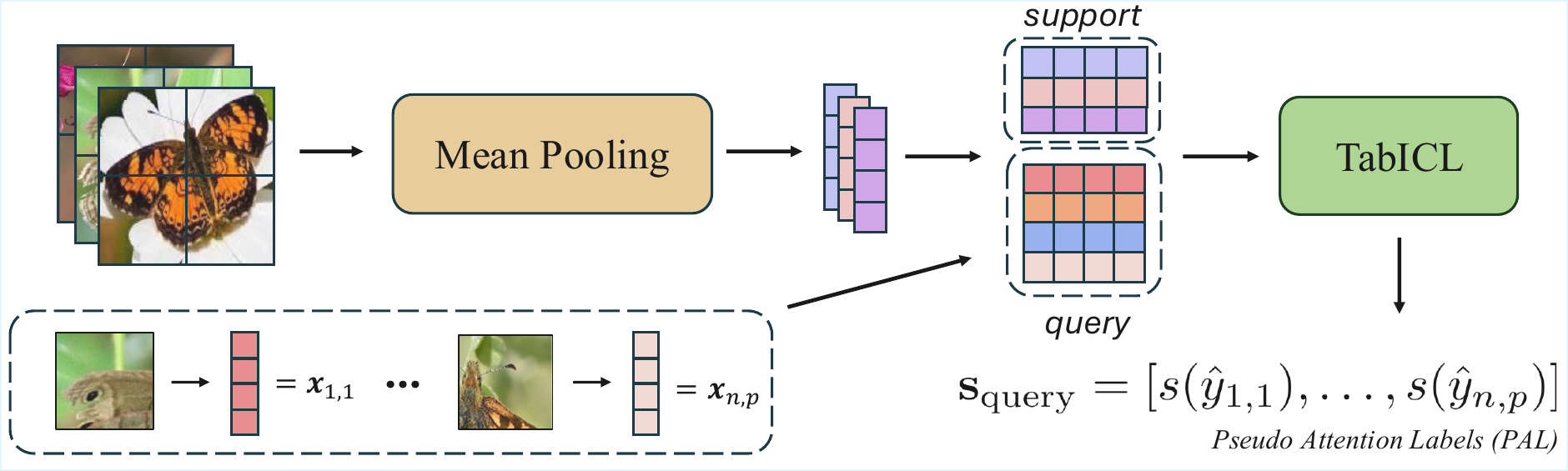}
    \caption{Construction of Pseudo Attention Labels. The support set comprises pooled text or image representations, whereas the query comprises more fine-grained tokens. The final PALs are produced by applying a scoring function, $s(\cdot)$, to the token-level predicted probability distributions.}
    \label{fig:pal_pooling}
\end{figure}

\begin{algorithm}[t]
\caption{PALPooler Fitting}
\label{alg:pal_pooling}
\begin{algorithmic}[1]
\Require Training dataset $\mathcal{D}^{\text{tok}}_{\text{train}}$, scoring function $s(\cdot)$, max queries $q_{\max}$, l2 reg $\lambda$, temperature $\tau$
\State Initialize parameters $\boldsymbol{\hat{\theta}} = \mathbf{0} \in \mathbb{R}^d$ \Comment{ {\footnotesize \textit{Yields mean-pooling via softmax}}}
\State Define weight generation: $\mathbf{w}\left(\mathbf{x}_i; \boldsymbol{\hat{\theta}}\right) := \mathrm{softmax}\left(\mathbf{X}_i \boldsymbol{\hat{\theta}}\right) \in \mathbb{R}^P$
\State Define pooling function: $\mathrm{Pool} \left(\mathbf{X}_i; \boldsymbol{\hat{\theta}} \right) := \sum_{p=1}^P \mathbf{w}_p\left(\mathbf{X}_i; \boldsymbol{\hat{\theta}} \right) \cdot \mathbf{x}_{i,p}$
\For{$t = 1, \dots, 3$}
    \State $\mathcal{D}_{\text{sup}}^{\text{tok}}, \mathcal{D}_{\text{qry}}^{\text{tok}} \gets \mathrm{RandomSplit}(\mathcal{D}_{\text{train}}, 0.5)$
    \State $\mathcal{D}_{\text{sup}} \gets \left\{ \left(\mathrm{Pool}(\mathbf{X}_i; \boldsymbol{\hat{\theta} }), y_i\right) \right\}_{i \in \mathcal{I}_{\text{sup}}}$ \Comment{ {\footnotesize \textit{Apply current pooler to support:} $\mathbb{R}^{\frac{n}{2} \times P \times d} \rightarrow \mathbb{R}^{\frac{n}{2} \times d}$ }}
    \State $\mathbf{Z}_{\text{qry}} \gets \mathrm{Sample} \left(\mathrm{Flatten}(\mathcal{D}_{\text{qry}}^{\text{tok}}), q_{\text{max}} \right)$ \Comment{{\footnotesize \textit{Discard labels and flatten:} $\mathbb{R}^{\frac{n}{2} \times P \times d} \rightarrow \mathbb{R}^{q_{\text{max}} \times d}$}}
    \State $\hat{\mathbf{Y}}_{\text{qry}} \gets \mathrm{TFM}\left(\mathbf{Z}_{\text{qry}}, \mathcal{D}_{\text{sup}} \right)$
    \State $\mathbf{s}_{\text{qry}} \gets s \left(\hat{\mathbf{Y}}_{\text{qry}} \right) / \tau$
    \State $\boldsymbol{\hat{\theta}} \gets \left(\mathbf{Z}_{\text{qry}}^\top \mathbf{Z}_{\text{qry}} + \lambda \mathbf{I} \right)^{-1} \mathbf{Z}_{\text{qry}}^\top \mathbf{s}_{\text{qry}}$ \Comment{ {\footnotesize \textit{Closed-form Ridge Regression}}}
\EndFor
\State \Return Learned pooling parameters $\boldsymbol{\hat{\theta}}$
\end{algorithmic}
\end{algorithm}

Formally, our goal is to learn a pooling function $\mathrm{Pool}(\cdot)$, parameterized by a linear scoring vector $\boldsymbol{\theta} \in \R^d$, which outputs a weighted average of token representations. To fit this model, we split the modality specific training set, $\mathcal{D}^{m, \text{tok}}_{\text{train}}$, into two distinct sets, pooling the first, $\mathcal{D}^{m, \text{tok}}_{\text{sup}}$, to construct our support and using the second, $\mathcal{D}^{m, \text{tok}}_{\text{qry}}$, as the query set of individual tokens for which we will derive Pseudo Attention Labels (PALs). These labels are constructed through a scoring function $s(\cdot): [0,1]^{C} \rightarrow \R$, that takes as input the predicted label distribution. For convenience, we drop the modality superscript $m$ as the method is only applied to a single modality at a time. The flow is illustrated in Figure \ref{fig:pal_pooling}, the fitting is described in Algorithm \ref{alg:pal_pooling}\footnote{Note that in practice, we always fit PCA on the pooled support, but exclude this in the formalization for clarity.}, and we discuss the scoring function $s$ below. As demonstrated in Appendix \ref{sec:PALPooling}, running PALPooling more than once, using the previously fitted pooler to refine the support, yields a positive impact; we therefore run this method iteratively. 
 
Choosing the scoring function $s$ is a crucial design choice for this method. We find that naive approaches, such as basing the score on the correct-class prediction probability or prediction entropy perform poorly on non-trivial datasets (see Figure \ref{fig:bad_scoring_functions} in Appendix). Instead, we find that the most stable solutions measure the distributional distance between the predictions and the prior label frequency in the dataset, motivated by the idea that tokens that move the prediction from the prior distribution are \textit{generally} more informative. That is, let $\mathbf{\bar{y}} \in [0,1]^{C}$ be the empirical label frequency, then $s(\mathbf{\hat{y}}_{i,p}) = \ln  \left(\mathrm{dist}(\mathbf{\hat{y}}_{i,p}, \mathbf{\bar{y}})\right)$, where $\mathrm{dist}$ is some function measuring distributional distance/divergence. In practice, we find the Jensen-Shannon Divergence (JSD)~\citep{lin2002divergence} to work well due to its bounded nature. In addition to fitting a pooling layer in seconds, PALPooling can be employed in settings where backpropagation through the TFM is impossible.

\section{Bridging TFMs and Non-Tabular Representations through PCA}

\label{sec:PCA}
\begin{figure}[t]
    \centering
    \includegraphics[width=\linewidth]{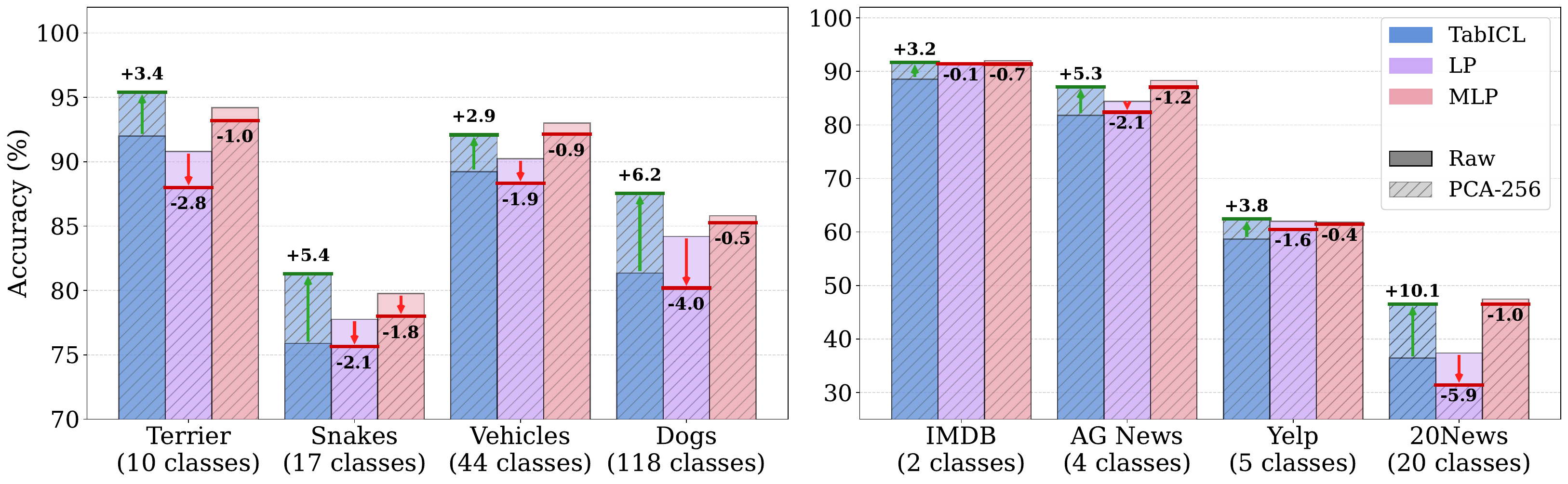}
    \caption{TabICL, Linear Probing (LP) and a 2-layer MLP with a ReLU activation evaluated on different datasets in text and image modalities with/without PCA. \textbf{Left:} Four subsets of ImageNet (described in Appendix \ref{sec:imagenetsubsets}). \textbf{Right:} Four text classification datasets. Across both modalities, PCA substantially improves TabICL's performance, enabling it to match or outperform the baselines.}
    \label{fig:cls_head}
\end{figure}

\begin{figure}[h]
    \centering
    \includegraphics[width=0.99\linewidth]{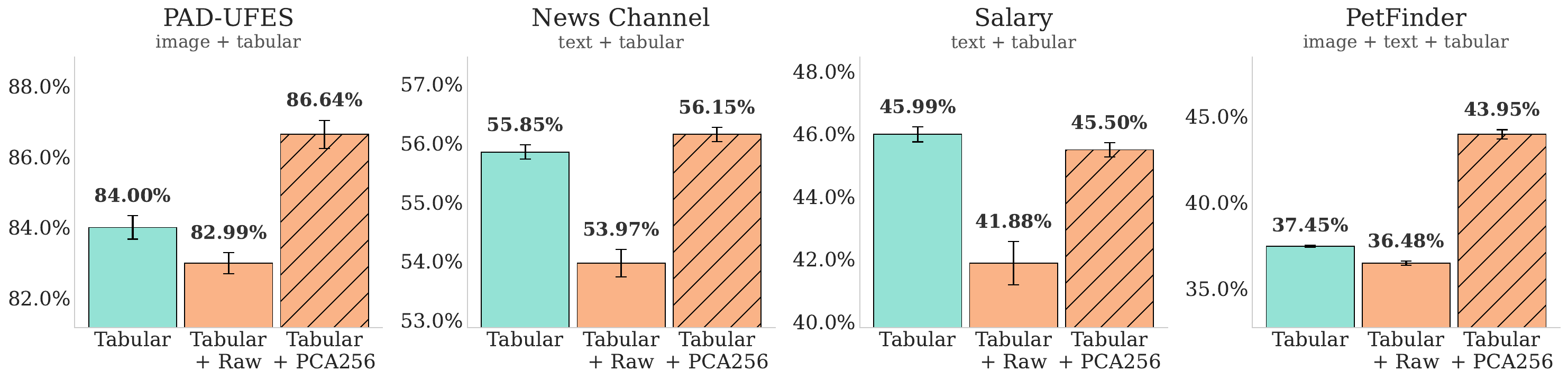}
    \caption{Accuracy on 4 datasets, where naively concatenating embeddings with tabular features deteriorates performance.}
    \label{fig:signal_balancing}
\end{figure}

We start by demonstrating the performance of TabICLv2~\citep{qu2026tabiclv2} as a general-purpose prediction head. TabICL is a suitable choice as it significantly alleviates many of the computational scaling issues of its peers. We perform experiments on image and text datasets, extracting image features using the \texttt{CLS}-token from DINOv3~\citep{simeoni2025dinov3}, and text features through ELECTRA~\citep{clark2020electra} with mean-pooling. Complete descriptions of the datasets can be found in Appendix \ref{sec:datasets}. In addition to comparing TabICL with classic probing baselines, we evaluate the performance of all methods after first reducing the embedding dimensionality to 256 through PCA. In this paper, we show that the benefits of PCA are two-fold:

\begin{enumerate}
    \item \textbf{Applying PCA has a significant positive effect on TabICL's image-only and text-only performance}. Figure \ref{fig:cls_head} compares the performance of TabICL against linear probing and a two-layer MLP on image-only and text-only classification. By applying PCA, TabICL achieves strong performance without any fine-tuning, often significantly outperforming linear probing. Importantly, the benefits of this projection do not extend to other classical probing heads, highlighting that TabICL's improvements stem from changes in the geometry of the data rather than information content. Text and image representations are generally of low effective rank~\citep{yin2018dimensionality}, and applying PCA significantly increases this. We investigate the relationship between this and TabICL's prior in Appendix \ref{sec:comparison}.
    \item \textbf{Applying PCA to extracted representations helps balance signals from tabular and non-tabular features.} In Figure \ref{fig:signal_balancing}, we run experiments on four datasets where the tabular features hold considerable signal: News Channel, Salary, PAD-UFES, and Petfinder. We find that PCA is crucial to avoid performance degradation in these settings.
\end{enumerate}

The two benefits discussed here motivate why PCA is a crucial component in the CoMET setup. In Appendix \ref{sec:tabpfn_discussion}, we include results for TabPFNv2.5~\citep{grinsztajn2025tabpfn}, with and without PCA. We note that, while it achieves competitive performance, it becomes computationally expensive to run for larger datasets, especially in the many-class setting.

\section{Experiments}

\subsection{Experimental Setup}

To emphasize the simplicity of CoMET, we do not perform any hyperparameter tuning. Unless otherwise stated, we use ELECTRA ($\Phi_\text{text}$) with mean-pooling, the \texttt{CLS}-token from DINOv3 ($\Phi_\text{img}$), and a PCA dimension of 256 (see Appendix~\ref{sec:pca_dim_choice} for an ablation). Additional information on all datasets used in the following experiments is provided in Appendix~\ref{sec:datasets}.

\paragraph{PALPooling} We first evaluate PALPooling in a unimodal setup, on both text and image datasets. Since texts vary in length, we employ a sample importance weight $w_{i} = (\sqrt{\mathrm{\#tokens}_i})^{-1}$ when fitting the ridge model to avoid overfitting the pooler to longer sequences. For $\Phi_\text{img}$, the number of individual tokens is fixed. Still, the spatial structure of images allows us to do more coarse-grained pooling by first aggregating individual tokens in groups of, for example, $2\times2$ or $4\times4$ using local mean-pooling. We observe that sequentially increasing the granularity leads to stable convergence, while reducing the number of tokens processed in earlier iterations, and, as such, employ group sizes $4\times4 \rightarrow 2\times2 \rightarrow 1\times1$ (see Appendix \ref{sec:PALPooling} for ablations). We run 3 iterations of pooling, with $q_{max} = 5 \times 10^5$, $\lambda = 10^4$, and PCA dimension 128 for all experiments, with $\tau = 0.5$ for image pooling and $\tau = 1$ for text pooling. For images, we run experiments on 4 datasets where we expect the \texttt{CLS}-token to struggle due to the presence of multiple semantically meaningful objects/features in a given image, and only a subset of these are important for the classification. These are: Pneumonia~\citep{shih2019augmenting}, Butterflies~\citep{depie2024}, MS COCO~\citep{lin2014microsoft}, and Open Images~\citep{kuznetsova2020open}. For texts, we evaluate on 4 popular datasets with varying text lengths, these are: IMDB Reviews~\citep{maas2011learning}, 20 Newsgroups~\citep{lang1995newsweeder}, Yelp Reviews~\citep{yelp2015dataset}, and AG News~\citep{zhang2015character}. For robustness, we include results on a collection of ImageNet tasks where the main object is easy to discern in Appendix \ref{sec:add_res}. Here, we expect the \texttt{CLS} token to be strong, but show that PALPooling consistently outperforms mean pooling, the starting point of the algorithm.

\begin{table}[h]
  \centering
  \caption{Dataset statistics of the datasets used for out multimodal results. Max seq.\ len is capped at the maximum ELECTRA token sequence length of 512. We subsample Jigsaw to 100,000/212,115 samples for train and 25,000/53,089 samples for test due to computational constraints.}
  \label{tab:mm-datasets}
  \begin{tabular}{lrrrcccc}
    \toprule
    \textbf{Dataset} & \textbf{Train} & \textbf{Test} & \textbf{Tab.\ feats} & \textbf{Max seq.\ len} & \textbf{Image} & \textbf{Text} & \textbf{Tabular} \\
    \midrule
    WikiArt         &  57{,}010 & 24{,}434 &  2 & ---  & \checkmark &            & \checkmark \\
    PetFinder       &   8{,}791 &  5{,}861 &  9 & 512  & \checkmark & \checkmark & \checkmark \\
    MM-IMDb         &  18{,}171 &  7{,}788 &  0 & 512  & \checkmark & \checkmark &            \\
    Airbnb          &  15{,}794 &  6{,}769 & 50 & 512  &            & \checkmark & \checkmark \\
    Wine Reviews    &  84{,}123 & 21{,}031 &  4 & 194  &            & \checkmark & \checkmark \\
    Jigsaw Toxicity & 100{,}000 & 25{,}000 & 29 & 358  &            & \checkmark & \checkmark \\
    \bottomrule
  \end{tabular}
\end{table}

\paragraph{Multimodal Evaluation} We evaluate multimodal classification performance on datasets that mix tabular, text, and image data. An overview of the datasets is given in Table \ref{tab:mm-datasets}. We compare the performance of our compositional setup against standard baselines as well as SOTA multimodal frameworks. For \textit{Text + Tabular} baselines we include TTT~\citep{bonnier2024revisiting} and TabSTAR~\citep{arazi2025tabstar},  trained models that operate on raw tokens rather than mean-pooled ELECTRA embeddings. For \textit{Image + Tabular} and \textit{Image + Text + Tabular}, we include MMPFN~\citep{kim2026multimodalpfn}, AutoGluon~\citep{tang2024autogluon} and Catboost~\citep{prokhorenkova2018catboost}. Recent work \citet{ye2025closerlooktabpfnv2} and \citet{luo2025time} has shown that TFMs can serve as powerful feature encoders of tabular data. Hence, we also include a Tabular Encoder (TE) baseline that extracts tabular embeddings using TabICL, concatenates them with the text/image embeddings, and then feeds the result into a 2-layer MLP. The tabular embeddings are constructed using the leave-one-out method from \cite{ye2025closerlooktabpfnv2}.

\subsection{Results}
\label{sec:results}

\begin{table*}[t]
  \centering
  \small
  \setlength{\tabcolsep}{4pt}
  \caption{Unimodal PALPooling results on image (\textbf{left}) and text (\textbf{right}) datasets, averaged over 5 seeds. PAL accuracy is taken at the best-validation iteration; fit time is cumulative up to that iteration.}
  \label{tab:pal_pooling}
  \begin{subtable}[t]{0.48\textwidth}
    \centering
    \label{tab:pal_pooling_image}
    \begin{tabular}{llccc c}
      \toprule
      Dataset & $n_{train}$ & \texttt{CLS} & Mean & PAL & Time \\
      \midrule
      \multirow{2}{*}{Pneumonia} & 20\% & \makecell{82.04\\[-3pt]{\scriptsize (0.07)}} & \makecell{81.73\\[-3pt]{\scriptsize (0.23)}} & \makecell{\textbf{82.95}\\[-3pt]{\scriptsize (0.27)}} & 16s \\
      \addlinespace[1pt]
       & 100\% & \makecell{82.96\\[-3pt]{\scriptsize (0.02)}} & \makecell{82.18\\[-3pt]{\scriptsize (0.00)}} & \makecell{\textbf{83.73}\\[-3pt]{\scriptsize (0.24)}} & 48s \\
      \cmidrule{1-6}
      \multirow{2}{*}{Butterflies} & 20\% & \makecell{59.94\\[-3pt]{\scriptsize (1.67)}} & \makecell{79.80\\[-3pt]{\scriptsize (0.71)}} & \makecell{\textbf{84.68}\\[-3pt]{\scriptsize (1.72)}} & 8s \\
      \addlinespace[1pt]
       & 100\% & \makecell{76.80\\[-3pt]{\scriptsize (0.22)}} & \makecell{90.28\\[-3pt]{\scriptsize (0.32)}} & \makecell{\textbf{93.67}\\[-3pt]{\scriptsize (0.53)}} & 54s \\
      \cmidrule{1-6}
      \multirow{2}{*}{MS COCO} & 20\% & \makecell{88.95\\[-3pt]{\scriptsize (0.29)}} & \makecell{91.55\\[-3pt]{\scriptsize (0.35)}} & \makecell{\textbf{91.74}\\[-3pt]{\scriptsize (0.24)}} & 36s \\
      \addlinespace[1pt]
       & 100\% & \makecell{91.60\\[-3pt]{\scriptsize (0.01)}} & \makecell{93.74\\[-3pt]{\scriptsize (0.00)}} & \makecell{\textbf{94.12}\\[-3pt]{\scriptsize (0.03)}} & 1.2m \\
      \cmidrule{1-6}
      \multirow{2}{*}{Open Images} & 20\% & \makecell{85.97\\[-3pt]{\scriptsize (0.15)}} & \makecell{91.08\\[-3pt]{\scriptsize (0.14)}} & \makecell{\textbf{91.59}\\[-3pt]{\scriptsize (0.24)}} & 43s \\
      \addlinespace[1pt]
       & 100\% & \makecell{88.69\\[-3pt]{\scriptsize (0.01)}} & \makecell{93.07\\[-3pt]{\scriptsize (0.01)}} & \makecell{\textbf{93.18}\\[-3pt]{\scriptsize (0.03)}} & 52s \\
      \bottomrule
    \end{tabular}
  \end{subtable}
  \hfill
  \begin{subtable}[t]{0.48\textwidth}
    \centering
    \label{tab:pal_pooling_text}
    \begin{tabular}{llccc c}
      \toprule
      Dataset & $n_{train}$ & \texttt{CLS} & Mean & PAL & Time \\
      \midrule
      \multirow{2}{*}{IMDB} & 20\% & \makecell{87.68\\[-3pt]{\scriptsize (0.19)}} & \makecell{90.05\\[-3pt]{\scriptsize (0.21)}} & \makecell{\textbf{90.23}\\[-3pt]{\scriptsize (1.07)}} & 1.0m \\
      \addlinespace[1pt]
       & 100\% & \makecell{89.10\\[-3pt]{\scriptsize (0.02)}} & \makecell{91.02\\[-3pt]{\scriptsize (0.00)}} & \makecell{\textbf{91.31}\\[-3pt]{\scriptsize (0.61)}} & 2.4m \\
      \cmidrule{1-6}
      \multirow{2}{*}{20 News} & 20\% & \makecell{28.09\\[-3pt]{\scriptsize (0.46)}} & \makecell{36.09\\[-3pt]{\scriptsize (0.58)}} & \makecell{\textbf{36.37}\\[-3pt]{\scriptsize (1.03)}} & 16s \\
      \addlinespace[1pt]
       & 100\% & \makecell{37.54\\[-3pt]{\scriptsize (0.02)}} & \makecell{45.44\\[-3pt]{\scriptsize (0.00)}} & \makecell{\textbf{46.67}\\[-3pt]{\scriptsize (0.99)}} & 1.7m \\
      \cmidrule{1-6}
      \multirow{2}{*}{Yelp} & 20\% & \makecell{66.02\\[-3pt]{\scriptsize (0.10)}} & \makecell{68.07\\[-3pt]{\scriptsize (0.11)}} & \makecell{\textbf{68.38}\\[-3pt]{\scriptsize (0.42)}} & 1.0m \\
      \addlinespace[1pt]
       & 100\% & \makecell{67.39\\[-3pt]{\scriptsize (0.00)}} & \makecell{69.17\\[-3pt]{\scriptsize (0.00)}} & \makecell{\textbf{69.81}\\[-3pt]{\scriptsize (0.38)}} & 1.8m \\
      \cmidrule{1-6}
      \multirow{2}{*}{AG News} & 20\% & \makecell{78.89\\[-3pt]{\scriptsize (0.19)}} & \makecell{83.20\\[-3pt]{\scriptsize (0.15)}} & \makecell{\textbf{84.04}\\[-3pt]{\scriptsize (0.33)}} & 46s \\
      \addlinespace[1pt]
       & 100\% & \makecell{81.17\\[-3pt]{\scriptsize (0.01)}} & \makecell{85.54\\[-3pt]{\scriptsize (0.00)}} & \makecell{\textbf{85.94}\\[-3pt]{\scriptsize (0.32)}} & 1.4m \\
      \bottomrule
    \end{tabular}
  \end{subtable}
\end{table*}

\begin{figure}[t]
  \centering

  \begin{subfigure}[b]{0.32\textwidth}
    \centering
    \setlength{\tabcolsep}{1pt}
    \renewcommand{\arraystretch}{0}
    \begin{tabular}{@{}cc@{}}
      \includegraphics[width=0.48\linewidth]{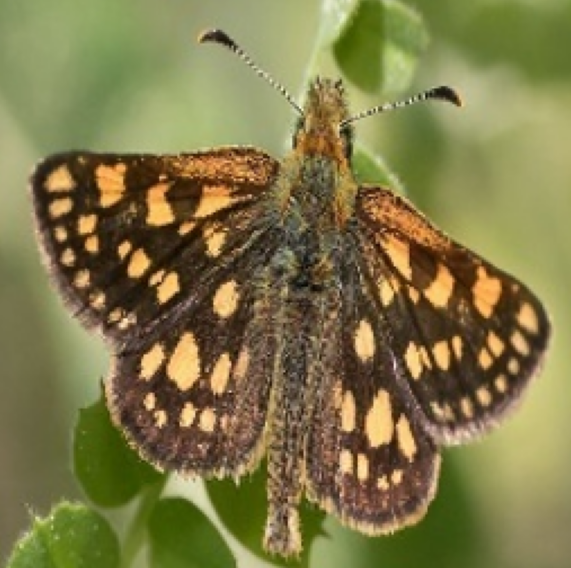} &
      \includegraphics[width=0.48\linewidth]{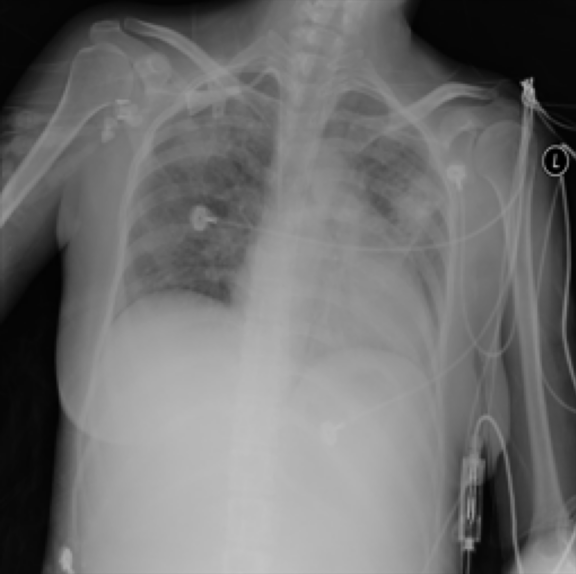} \\[1.5pt]
      \includegraphics[width=0.48\linewidth]{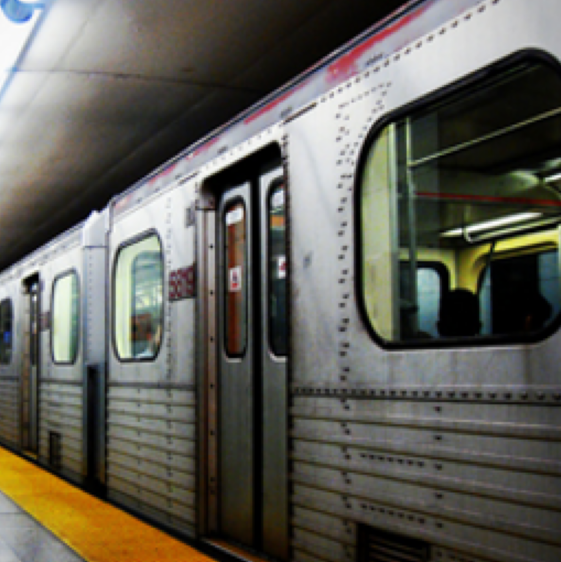} &
      \includegraphics[width=0.48\linewidth]{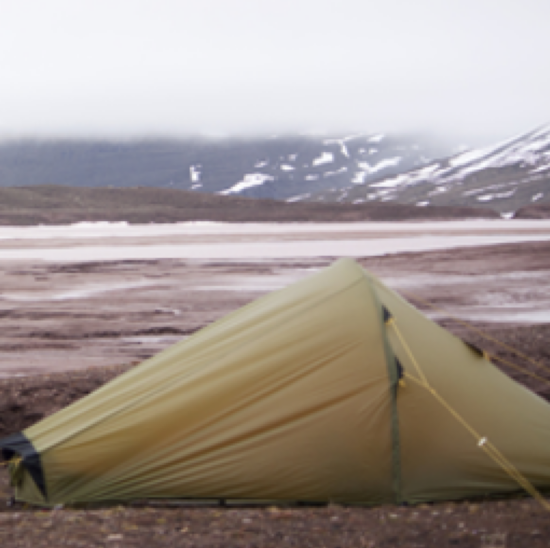}
    \end{tabular}
    \caption{Original}
  \end{subfigure}%
  \hfill
  \begin{subfigure}[b]{0.32\textwidth}
    \centering
    \setlength{\tabcolsep}{1pt}
    \renewcommand{\arraystretch}{0}
    \begin{tabular}{@{}cc@{}}
      \includegraphics[width=0.48\linewidth]{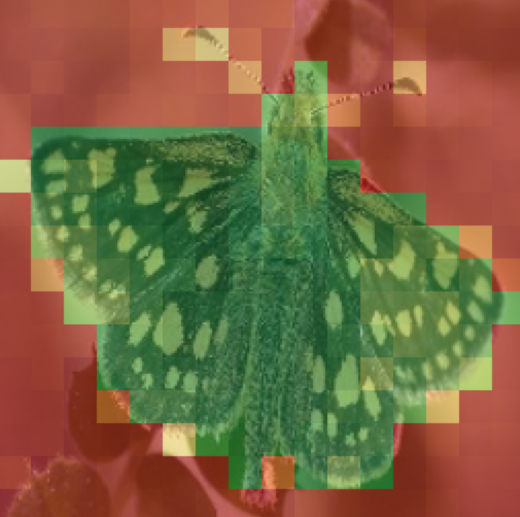} &
      \includegraphics[width=0.48\linewidth]{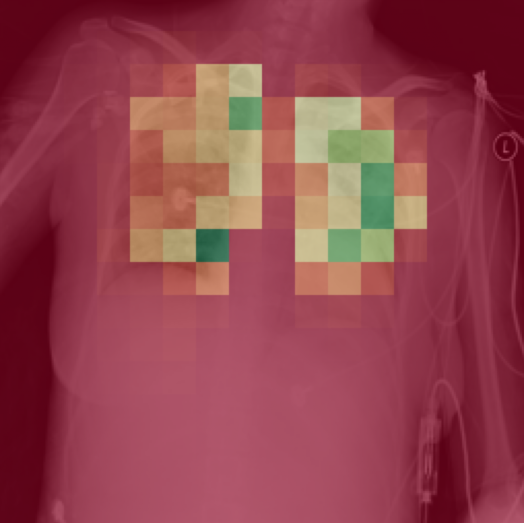} \\[1.5pt]
      \includegraphics[width=0.48\linewidth]{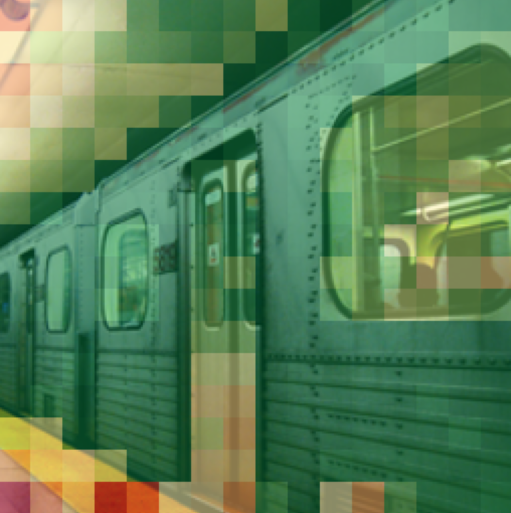} &
      \includegraphics[width=0.48\linewidth]{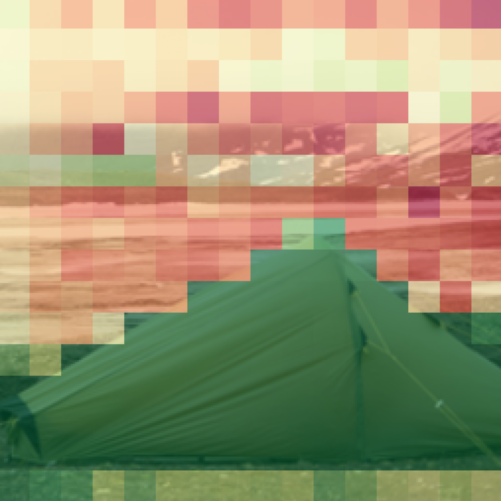}
    \end{tabular}
    \caption{PALs: $s(\hat{\mathbf{y}}_{i,p})$}
  \end{subfigure}%
  \hfill
  \begin{subfigure}[b]{0.32\textwidth}
    \centering
    \setlength{\tabcolsep}{1pt}
    \renewcommand{\arraystretch}{0}
    \begin{tabular}{@{}cc@{}}
      \includegraphics[width=0.48\linewidth]{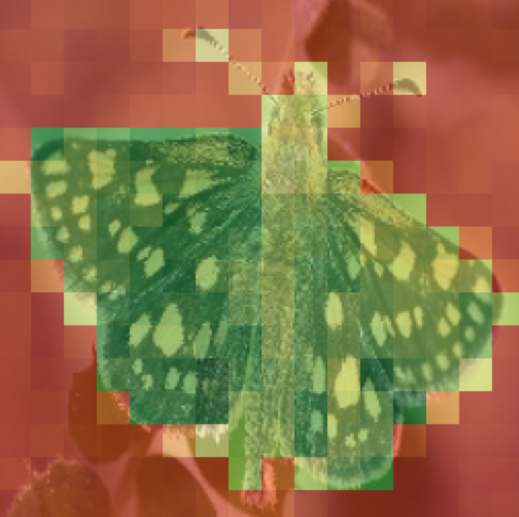} &
      \includegraphics[width=0.48\linewidth]{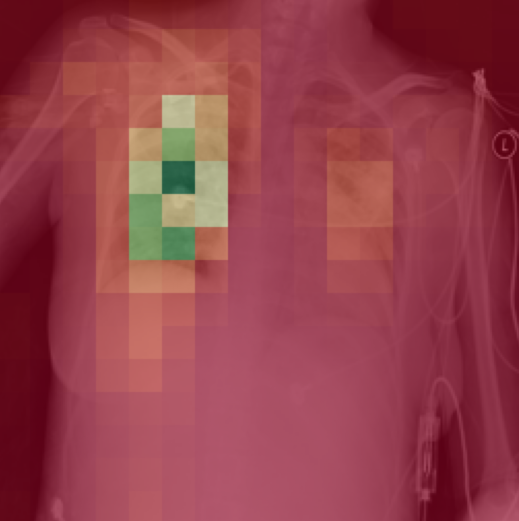} \\[1.5pt]
      \includegraphics[width=0.48\linewidth]{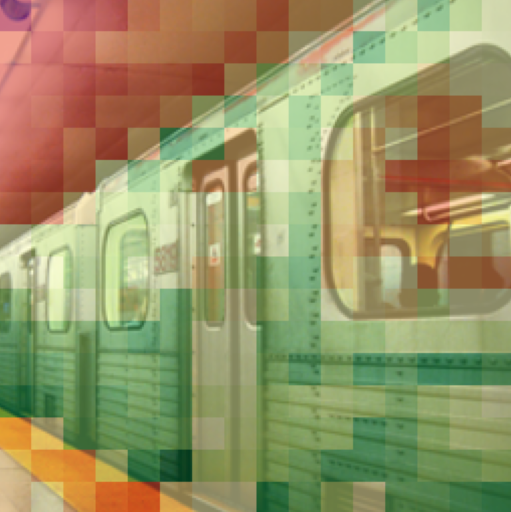} &
      \includegraphics[width=0.48\linewidth]{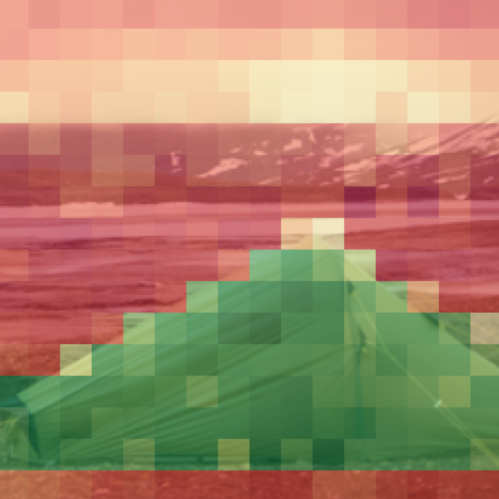}
    \end{tabular}
    \caption{Pooling Weights: $\boldsymbol{\hat{\theta}}^{\top}\mathbf{x}_{i,p}$}
  \end{subfigure}
  \caption{Example of PALs and the final pooling weights from the image datasets in Table \ref{tab:pal_pooling}. Top-left: Butterflies. Top-right: Pneumonia. Bottom-left: MS COCO. Bottom-right: Open Images.}
  \label{fig:pal_viz}
\end{figure}

The results of the unimodal PALPooling experiments are shown in Table \ref{tab:pal_pooling}. While the magnitude of the performance gain varies across datasets and sizes, we observe that it can significantly increase accuracy with only a few seconds of additional overhead, with improvements most pronounced on Pneumonia and Butterflies. Figure \ref{fig:pal_viz} shows visualizations of the produced pseudo attention labels and the final learned pooling model. The Pneumonia results demonstrate that PALs can go beyond simple foreground-background separation, capturing finer-grained details such as the lung regions.

Table~\ref{tab:tabicl_results} reports the results of the multimodal experiments. Across nearly all datasets, CoMET matches or beats baselines, including MMPFN, the closest prior work that leverages TFMs for multimodal prediction. The notable exception is the Wine dataset, where TTT and TabSTAR outperform our approach. Both of these methods specialize in tabular+text data and operate at a finer granularity, consuming individual token embeddings directly instead of an aggregate, which could be beneficial for some tasks. The performance difference between CoMET and the TE method highlights the added benefit of not using TFMs as tabular-only encoders; instead, CoMET leverages their expressivity to enable richer cross-modal interactions.

\begin{table*}[t]
\centering
\caption{
Accuracy on multimodal datasets over 5 seeds. Best result per dataset is shown in \textbf{bold}. PF = PetFinder. For WikiArt and Wine, we randomly subsample to 10 classes per seed since MM-PFN only supports datasets with at most 10 classes (see Appendix~\ref{sec:mmtabicl_discussion}). T=Text, t=Tabular and I=Image.}
\label{tab:tabicl_results}
\resizebox{\textwidth}{!}{%
\begin{tabular}{llcccccccc}
\toprule
& \textbf{Dataset}
& \shortstack{\textbf{CoMET} \\ \textbf{+ PAL}}
& \textbf{CoMET}
& \textbf{TE}
& \shortstack{\textbf{Auto} \\ \textbf{Gluon}}
& \shortstack{\textbf{Cat} \\ \textbf{Boost}}
& \shortstack{\textbf{MM-} \\ \textbf{PFN}}
& \textbf{TTT}
& \shortstack{\textbf{Tab-} \\ \textbf{STAR}} \\
\midrule
\textit{T + t}
& Jigsaw
& 95.25 & 94.72 & 94.74 & 94.73 & 94.75 & 94.21 & 95.68 & \textbf{95.94} \\[-0.5ex]
&
& {\scriptsize (0.03)} & {\scriptsize (0.04)} & {\scriptsize (0.02)}
& {\scriptsize (0.03)} & {\scriptsize (0.03)} & {\scriptsize (0.01)}
& {\scriptsize (0.08)} & \textbf{\scriptsize (0.14)} \\
\addlinespace[4pt]
&
Wine
& 83.18 & 82.54 & 78.85 & 79.89 & 78.04 & 71.14 & 88.80 & \textbf{90.79} \\[-0.5ex]
&
& {\scriptsize (1.85)} & {\scriptsize (1.81)} & {\scriptsize (2.04)}
& {\scriptsize (1.84)} & {\scriptsize (2.05)} & {\scriptsize (1.73)}
& {\scriptsize (2.81)} & \textbf{\scriptsize (2.63)} \\
\addlinespace[4pt]
&
Airbnb
& 45.15 & \textbf{45.17} & 43.87 & 39.89 & 38.97 & 43.12 & 37.57 & 40.52 \\[-0.5ex]
&
& {\scriptsize (0.23)} & \textbf{\scriptsize (0.24)} & {\scriptsize (0.36)}
& {\scriptsize (0.51)} & {\scriptsize (0.23)} & {\scriptsize (0.21)}
& {\scriptsize (0.48)} & {\scriptsize (0.33)} \\
\addlinespace[4pt]
&
PF-Text
& 42.78 & \textbf{43.26} & 36.49 & 40.53 & 39.20 & 38.29 & 36.46 & 38.19 \\[-0.5ex]
&
& {\scriptsize (0.07)} & \textbf{\scriptsize (0.17)} & {\scriptsize (0.08)}
& {\scriptsize (0.19)} & {\scriptsize (0.35)} & {\scriptsize (0.22)}
& {\scriptsize (0.31)} & {\scriptsize (0.50)} \\
\midrule
\textit{I + t}
& WikiArt
& \textbf{89.04} & 89.01 & 84.49 & 82.95 & 82.59 & 81.42 & $-$ & $-$ \\[-0.5ex]
&
& \textbf{\scriptsize (1.17)} & {\scriptsize (1.23)} & {\scriptsize (1.81)}
& {\scriptsize (1.69)} & {\scriptsize (1.57)} & {\scriptsize (0.45)}
& & \\
\addlinespace[4pt]
&
PF-Img
& 40.46 & \textbf{41.22} & 38.05 & 39.54 & 37.50 & 39.94 & $-$ & $-$ \\[-0.5ex]
&
& {\scriptsize (0.26)} & \textbf{\scriptsize (0.15)} & {\scriptsize (0.13)}
& {\scriptsize (0.01)} & {\scriptsize (0.22)} & {\scriptsize (0.21)}
& & \\
\midrule
\textit{I + T + t}
& PF-All
& 42.60 & \textbf{44.38} & 38.14 & 41.68 & 40.14 & 40.48 & $-$ & $-$ \\[-0.5ex]
&
& {\scriptsize (0.29)} & \textbf{\scriptsize (0.24)} & {\scriptsize (0.19)}
& {\scriptsize (0.00)} & {\scriptsize (0.48)} & {\scriptsize (0.28)}
& & \\
\midrule
\textit{I + T}
& MM-IMDb
& \textbf{63.95} & 63.62 & 54.89 & 63.52 & 62.79 & $-$ & $-$ & $-$ \\[-0.5ex]
&
& \textbf{\scriptsize (0.37)} & {\scriptsize (0.20)} & {\scriptsize (0.01)}
& {\scriptsize (0.20)} & {\scriptsize (0.34)}
& & & \\
\bottomrule
\end{tabular}%
}
\end{table*}

The benefits of PALPooling vary across multimodal datasets, offering a boost on Jigsaw, Wine, and MM-IMDB, performing similarly on Airbnb, and performing worse on Petfinder. As previously discussed, we do not expect PALPooling to always outperform the \texttt{CLS} token, and its effectiveness will depend on the extent to which the \texttt{CLS} and mean-pooled representations fail to capture the signal. Furthermore, because the pooler is agnostic to features from other modalities, it is not necessarily emphasizing complementary information (see Figure \ref{fig:petfinder_cls_pal} in Appendix \ref{sec:add_res} for example).

\section{Fast and Accurate Predictions on Datasets with Hierarchical Labels}
\label{sec:hier}

\begin{wrapfigure}{r}{0.48\linewidth}
    \centering
    \includegraphics[width=\linewidth]{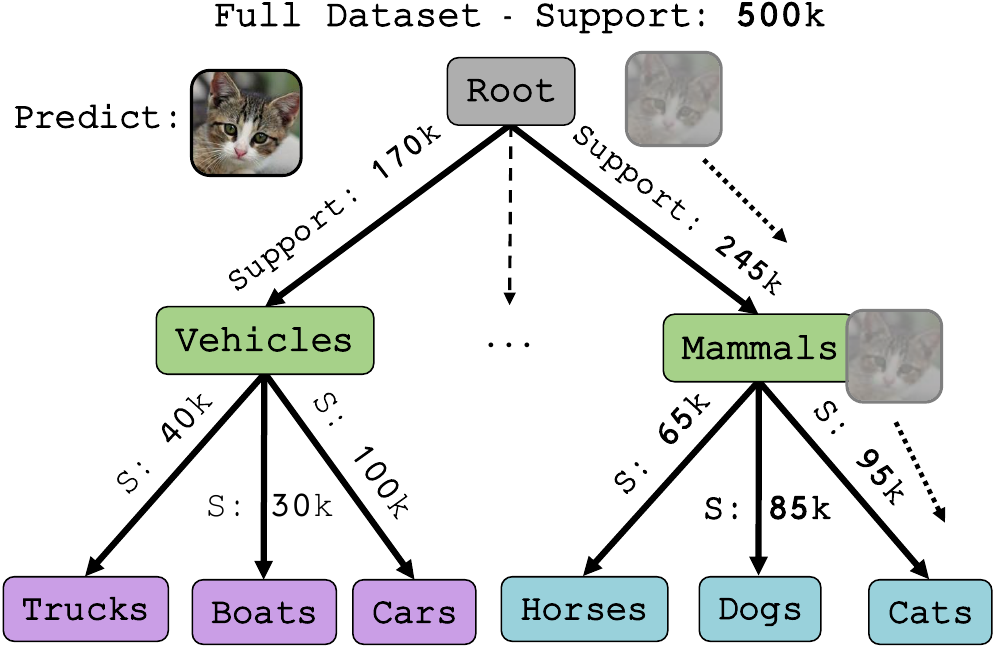}
    \caption{Hierarchical classification with CoMET. Given a broad classification problem, we construct a hierarchical tree in which the target classes reside at the leaf nodes and are grouped into higher-level categories represented by parent nodes. Each node defines a sub-classification problem over its children. We leverage TabICLv2 for each subproblem, constructing a fine-grained support by only including training samples from that category, and making the final predictions by traversing the hierarchy from root to leaf node, at each step forwarding the evaluation sample to the predicted node.}
    \label{fig:hier}
\end{wrapfigure}

The setting of hierarchical classification, in which classes can be grouped into subcategories, is common in data across various modalities~\citep{lewis2004rcv1, bang2023gacaps, jiang2022exploiting, park2024visually}. Given the robust performance of CoMET, we now explore this as an interesting application where strong training-free predictions can be readily leveraged.  We argue that the use of TFMs is motivated in this setting for two reasons: Firstly, when predicting subcategories further down the hierarchy, it is possible to remove uninteresting samples from the support sets, making them increasingly tailored to their specific task. Secondly, since TFMs are training-free, this so-called local approach~\citep{zangari2024hierarchical} can be leveraged without training separate models for each subtask, making it a feasible and computationally efficient approach for large hierarchies. We refer to this hierarchical inference model as H-CoMET, and describe it in Figure \ref{fig:hier}. We compare this method to Flat-CoMET (F-CoMET) and a 2-layer MLP (F-MLP), both of which predict the leaf nodes directly, without using a hierarchy. For computational feasibility, F-CoMET and H-CoMET have a max support budget of 200,000 for any problem/subproblem. F-MLP uses 90\% of the \emph{full} training set for fitting, and 10\% for model selection. 

\paragraph{Evaluation} We evaluate the performance of our hierarchical approach on a combination of text, image, and tabular data. Specifically, to first highlight the competitiveness of H-CoMET, we evaluate on the text benchmarks and baseline results provided by \citet{zangari2024hierarchical}\footnote{Excluding multilabel and non-
mandatory leaf node prediction datasets.}. The datasets include Amazon Reviews~\citep{hou2024bridging}, Web of Science (WOS)~\citep{kowsari2017HDLTex} (CC BY 4.0), and Linux Bugs~\citep{lyubinets2018automated} (CC BY 4.0), and the baselines MATCH~\citep{zhang2021match}, HiAGM~\citep{zhou2020hierarchy}, HBGL~\citep{jiang2022exploiting}, GACaps~\citep{bang2023gacaps}, and a standard fine-tuned flat BERT model~\citep{devlin2019bert}. We refer to \citet{zangari2024hierarchical} for a thorough description of these. While the baselines are often fine-tuned end-to-end, we instead rely on a fixed backbone and pooling method for each dataset. More precisely, we use Sentence-BERT~\citep{reimers-2019-sentence-bert} for Amazon Reviews and CodeBERT~\citep{feng2020codebert} for Bugs, and aggregate both using mean-pooling. Given its strong performance in \citet{zangari2024hierarchical}, we use TF–IDF~\citep{sparck1972statistical} followed by chi-squared feature selection for WOS, resulting in a similar dimensionality of 768. Since CoMET is a multimodal method, we additionally derive two datasets of our own, with considerably vaster hierarchies, from INaturalist \cite{vanhorn2018inaturalistspeciesclassificationdetection} (image + tabular), and a larger version of Amazon Reviews (Amazon MM), which includes more product features (image + text + tabular). We refer to Appendix \ref{sec:preprocess_hier} for details. Table \ref{tab:hier_datasets} provides an overview of the dataset and the average execution time of H-CoMET, showing that performing inference with a TabICL classifier per-node finishes in a reasonable amount of time, even when the number of subtasks exceeds 1000.

\begin{table}[t]
\centering
\caption{Overview of the datasets used. Subtasks indicate the number of distinct TabICL supports used. Time refers to fitting + inference time for H-CoMET }
\label{tab:hier_datasets}
\begin{tabular}{lccccccc}
\toprule
Dataset & $n_{\text{train}}$ & $n_{\text{test}}$ & $n_{\text{classes}}$ & Depth & Subtasks & Time \\
\midrule
WOS          & 31k & 16k & 138 & 2 & 8 &  29s \\
Amazon       & 333k & 167k & 25 & 2 & 6 &  98s \\
Bugs         & 23k & 12k & 85 & 2 & 18 &  18s \\
iNaturalist  &  800k & 28k &  2895 & 7 & 1158 & 315s  \\
Amazon MM    & 444k & 55k & 3047 & 4 & 633 & 310s \\
\bottomrule
\end{tabular}
\end{table}

\paragraph{Results} The results in Table \ref{tab:hier_text} emphasize the benefit of the hierarchical approach, with H-CoMET consistently outperforming F-CoMET, and reaching performance comparable to state-of-the-art hierarchical text models. The improvements are particularly pronounced on the Bugs dataset, with H-CoMET achieving significantly higher accuracy than baselines, owing to its ability to better handle the fine-grained subcategory classification (illustrated in Appendix \ref{sec:text_benchmarks}). H-CoMET also scales well to the more extensive multimodal Amazon MM and iNaturalist hierarchies, achieving high accuracy using 5 minutes of fitting and inference time. The F-MLP baseline still outperforms H-CoMET on iNaturalist, likely because it can better leverage the full 800k samples. Still, these results suggest that, as TFMs develop and become more scalable, their ability to handle large and complex hierarchical structures with a limited context length and no additional training holds great promise.

\begin{table}[h]
\centering
\caption{Accuracy of H-CoMET, compared to our baselines and SOTA hierarchical text models.}
\label{tab:hier_text}
\begin{tabular}{lccccc}
\toprule
\textbf{Model} & \textbf{WOS} $\uparrow$ & \textbf{Amazon} $\uparrow$ & \textbf{Bugs} $\uparrow$ & \textbf{Amazon MM} $\uparrow$ & \textbf{iNaturalist} $\uparrow$ \\
\midrule
MATCH & 59.32 (1.61) & 87.17 (0.87) & 36.24 (4.91) & - & - \\
HiAGM & 65.13 (1.59) & 87.35 (0.44) & 44.82 (0.69) & - & - \\
HBGL & \textbf{80.14 (0.02)} & 84.05 (0.02) & 57.63 (0.03) & - & - \\
GACaps & 73.76 (0.78) & 86.73 (0.19) & 49.16 (0.52) & - & - \\
BERT & 77.12 (0.67) & 89.54 (0.20) & 50.70 (1.13) & - & - \\
\midrule
H-CoMET & 75.53 (0.38) & \textbf{89.85} (0.04) & \textbf{76.61} (0.44) & \textbf{78.55} (0.06) & 82.50 (0.04) \\
F-CoMET & 73.41 (0.28) & 87.17 (0.09) & 31.44 (0.41) & 73.19 (0.19) & 69.54 (0.23) \\
F-MLP & 76.16 (0.26) & 88.48 (0.07) & 31.40 (0.91) & 77.15 (0.08) & \textbf{83.81} (0.09) \\
\bottomrule
\end{tabular}
\end{table}

\section{Conclusion}

The results in this paper demonstrate the potential of composing strong modality encoders with in-context learners to make accurate predictions without task-specific fine-tuning. By balancing signals through PCA, CoMET achieves state-of-the-art performance on several multimodal benchmarks, despite the TFM never being explicitly trained on non-tabular data. As both TFMs and modality backbones improve, we anticipate that the gap between composing encoders with TFMs and fully trained end-to-end pipelines will continue to decrease. Longer context windows, larger class capacities, and richer pre-training priors may allow a single frozen compositional pipeline to ingest large databases of mixed-modality data and produce accurate predictions in seconds. More broadly, the modular nature of CoMET means it inherits future progress for free: new backbones or successor TFMs are drop-in upgrades, sidestepping the retraining and re-tuning that end-to-end pipelines typically require with each new release.

\textbf{Limitations and Future Work:} Our work is a first step, and several directions remain open. First, we relied on PCA as our adaptor between modality encoders and the TFM. While its simplicity is a start, more sophisticated TFM-informed dimensionality reduction methods could yield further gains without breaking the training-free pipeline. Furthermore, we focused on text and image modalities, but the same compositional template is general and extends naturally to any domain with strong off-the-shelf encoders, such as audio, video, and 3D. We have shown that our compositional setup benefits complex data structures such as hierarchies; combined with the rise of agentic workflows, we see a natural path toward CoMET-driven structure discovery, where agents iteratively propose and evaluate organizations' data without any model retraining.

\section*{Acknowledgments}

RGK gratefully acknowledges support from the Canada Research Chairs Program (CRC-2022-00049) and the Canada CIFAR AI Chairs Program, This research was funded in part by a NFRF Special Call Award (NFRFR-2022-00526) and NSERC Discovery Grant (RGPIN-2022-04546). HB is supported by Swedish Research Council Grant 2022-04748, as well as the Sweden-America Foundation. Resources used in preparing this research were provided, in part, by the Province of Ontario, the Government of Canada through CIFAR, and companies sponsoring the Vector Institute.

\newpage
\clearpage
\bibliographystyle{plainnat}
\bibliography{main}

\begin{thebibliography}{58}
\providecommand{\natexlab}[1]{#1}
\providecommand{\url}[1]{\texttt{#1}}
\expandafter\ifx\csname urlstyle\endcsname\relax
  \providecommand{\doi}[1]{doi: #1}\else
  \providecommand{\doi}{doi: \begingroup \urlstyle{rm}\Url}\fi

\bibitem[Arazi et~al.(2025)Arazi, Shapira, and Reichart]{arazi2025tabstar}
Alan Arazi, Eilam Shapira, and Roi Reichart.
\newblock Tabstar: A tabular foundation model for tabular data with text fields.
\newblock \emph{arXiv preprint arXiv:2505.18125}, 2025.

\bibitem[Arevalo et~al.(2017)Arevalo, Solorio, Montes-y G{\'o}mez, and Gonz{\'a}lez]{arevalo2017gated}
John Arevalo, Thamar Solorio, Manuel Montes-y G{\'o}mez, and Fabio~A. Gonz{\'a}lez.
\newblock Gated multimodal units for information fusion.
\newblock In \emph{5th International Conference on Learning Representations (ICLR) Workshop}, 2017.

\bibitem[Balazadeh et~al.(2025)Balazadeh, Kamkari, Thomas, Li, Ma, Cresswell, and Krishnan]{balazadeh2025causalpfn}
Vahid Balazadeh, Hamidreza Kamkari, Valentin Thomas, Benson Li, Junwei Ma, Jesse~C Cresswell, and Rahul~G Krishnan.
\newblock Causalpfn: Amortized causal effect estimation via in-context learning.
\newblock \emph{arXiv preprint arXiv:2506.07918}, 2025.

\bibitem[Bang et~al.(2023)Bang, Park, and Park]{bang2023gacaps}
Jinhyun Bang, Jonghun Park, and Jonghyuk Park.
\newblock Gacaps-htc: graph attention capsule network for hierarchical text classification.
\newblock \emph{Applied Intelligence}, 53\penalty0 (17):\penalty0 20577--20594, 2023.

\bibitem[Bonnier(2024)]{bonnier2024revisiting}
Thomas Bonnier.
\newblock Revisiting multimodal transformers for tabular data with text fields.
\newblock In \emph{Findings of the Association for Computational Linguistics: ACL 2024}, pages 1481--1500, 2024.

\bibitem[Clark et~al.(2020)Clark, Luong, Le, and Manning]{clark2020electra}
Kevin Clark, Minh-Thang Luong, Quoc~V Le, and Christopher~D Manning.
\newblock Electra: Pre-training text encoders as discriminators rather than generators.
\newblock \emph{arXiv preprint arXiv:2003.10555}, 2020.

\bibitem[Devlin et~al.(2019)Devlin, Chang, Lee, and Toutanova]{devlin2019bert}
Jacob Devlin, Ming-Wei Chang, Kenton Lee, and Kristina Toutanova.
\newblock Bert: Pre-training of deep bidirectional transformers for language understanding.
\newblock In \emph{Proceedings of the 2019 conference of the North American chapter of the association for computational linguistics: human language technologies, volume 1 (long and short papers)}, pages 4171--4186, 2019.

\bibitem[Dosovitskiy et~al.(2020)Dosovitskiy, Beyer, Kolesnikov, Weissenborn, Zhai, Unterthiner, Dehghani, Minderer, Heigold, Gelly, et~al.]{dosovitskiy2020image}
Alexey Dosovitskiy, Lucas Beyer, Alexander Kolesnikov, Dirk Weissenborn, Xiaohua Zhai, Thomas Unterthiner, Mostafa Dehghani, Matthias Minderer, Georg Heigold, Sylvain Gelly, et~al.
\newblock An image is worth 16x16 words: Transformers for image recognition at scale.
\newblock \emph{arXiv preprint arXiv:2010.11929}, 2020.

\bibitem[Du et~al.(2024)Du, Zheng, Wang, Bai, O’Regan, and Qin]{du2024tip}
Siyi Du, Shaoming Zheng, Yinsong Wang, Wenjia Bai, Declan~P O’Regan, and Chen Qin.
\newblock Tip: Tabular-image pre-training for multimodal classification with incomplete data.
\newblock In \emph{European Conference on Computer Vision}, pages 478--496. Springer, 2024.

\bibitem[Feng et~al.(2020)Feng, Guo, Tang, Duan, Feng, Gong, Shou, Qin, Liu, Jiang, et~al.]{feng2020codebert}
Zhangyin Feng, Daya Guo, Duyu Tang, Nan Duan, Xiaocheng Feng, Ming Gong, Linjun Shou, Bing Qin, Ting Liu, Daxin Jiang, et~al.
\newblock Codebert: A pre-trained model for programming and natural languages.
\newblock In \emph{Findings of the association for computational linguistics: EMNLP 2020}, pages 1536--1547, 2020.

\bibitem[Grinsztajn et~al.(2025)Grinsztajn, Fl{\"o}ge, Key, Birkel, Jund, Roof, J{\"a}ger, Safaric, Alessi, Hayler, et~al.]{grinsztajn2025tabpfn}
L{\'e}o Grinsztajn, Klemens Fl{\"o}ge, Oscar Key, Felix Birkel, Philipp Jund, Brendan Roof, Benjamin J{\"a}ger, Dominik Safaric, Simone Alessi, Adrian Hayler, et~al.
\newblock Tabpfn-2.5: Advancing the state of the art in tabular foundation models.
\newblock \emph{arXiv preprint arXiv:2511.08667}, 2025.

\bibitem[Hager et~al.(2023)Hager, Menten, and Rueckert]{Hager_2023_CVPR}
Paul Hager, Martin~J. Menten, and Daniel Rueckert.
\newblock Best of both worlds: Multimodal contrastive learning with tabular and imaging data.
\newblock In \emph{Proceedings of the IEEE/CVF Conference on Computer Vision and Pattern Recognition (CVPR)}, pages 23924--23935, June 2023.

\bibitem[Hollmann et~al.(2022)Hollmann, M{\"u}ller, Eggensperger, and Hutter]{hollmann2022tabpfn}
Noah Hollmann, Samuel M{\"u}ller, Katharina Eggensperger, and Frank Hutter.
\newblock Tabpfn: A transformer that solves small tabular classification problems in a second.
\newblock \emph{arXiv preprint arXiv:2207.01848}, 2022.

\bibitem[Hollmann et~al.(2025)Hollmann, M{\"u}ller, Purucker, Krishnakumar, K{\"o}rfer, Hoo, Schirrmeister, and Hutter]{hollmann2025accurate}
Noah Hollmann, Samuel M{\"u}ller, Lennart Purucker, Arjun Krishnakumar, Max K{\"o}rfer, Shi~Bin Hoo, Robin~Tibor Schirrmeister, and Frank Hutter.
\newblock Accurate predictions on small data with a tabular foundation model.
\newblock \emph{Nature}, 637\penalty0 (8045):\penalty0 319--326, 2025.

\bibitem[Horn et~al.(2018)Horn, Aodha, Song, Cui, Sun, Shepard, Adam, Perona, and Belongie]{vanhorn2018inaturalistspeciesclassificationdetection}
Grant~Van Horn, Oisin~Mac Aodha, Yang Song, Yin Cui, Chen Sun, Alex Shepard, Hartwig Adam, Pietro Perona, and Serge Belongie.
\newblock The inaturalist species classification and detection dataset, 2018.
\newblock URL \url{https://arxiv.org/abs/1707.06642}.

\bibitem[Hou et~al.(2024)Hou, Li, He, Yan, Chen, and McAuley]{hou2024bridging}
Yupeng Hou, Jiacheng Li, Zhankui He, An~Yan, Xiusi Chen, and Julian McAuley.
\newblock Bridging language and items for retrieval and recommendation.
\newblock \emph{arXiv preprint arXiv:2403.03952}, 2024.

\bibitem[Hu et~al.(2022)Hu, Shen, Wallis, Allen-Zhu, Li, Wang, Wang, Chen, et~al.]{hu2022lora}
Edward~J Hu, Yelong Shen, Phillip Wallis, Zeyuan Allen-Zhu, Yuanzhi Li, Shean Wang, Liang Wang, Weizhu Chen, et~al.
\newblock Lora: Low-rank adaptation of large language models.
\newblock \emph{Iclr}, 1\penalty0 (2):\penalty0 3, 2022.

\bibitem[Jiang et~al.(2022)Jiang, Wang, Sun, Chen, Zhuang, and Yang]{jiang2022exploiting}
Ting Jiang, Deqing Wang, Leilei Sun, Zhongzhi Chen, Fuzhen Zhuang, and Qinghong Yang.
\newblock Exploiting global and local hierarchies for hierarchical text classification.
\newblock In \emph{Proceedings of the 2022 conference on empirical methods in natural language processing}, pages 4030--4039, 2022.

\bibitem[{Jigsaw and Conversation AI}(2018)]{jigsaw2018toxic}
{Jigsaw and Conversation AI}.
\newblock Toxic comment classification challenge.
\newblock \url{https://www.kaggle.com/c/jigsaw-toxic-comment-classification-challenge}, 2018.

\bibitem[Kim et~al.(2026)Kim, Song, and Kim]{kim2026multimodalpfn}
Wall Kim, Chaeyoung Song, and Hanul Kim.
\newblock Multimodalpfn: Extending prior-data fitted networks for multimodal tabular learning.
\newblock \emph{arXiv preprint arXiv:2602.20223}, 2026.

\bibitem[Kowsari et~al.(2017)Kowsari, Brown, Heidarysafa, Jafari~Meimandi, , Gerber, and Barnes]{kowsari2017HDLTex}
Kamran Kowsari, Donald~E Brown, Mojtaba Heidarysafa, Kiana Jafari~Meimandi, , Matthew~S Gerber, and Laura~E Barnes.
\newblock Hdltex: Hierarchical deep learning for text classification.
\newblock In \emph{Machine Learning and Applications (ICMLA), 2017 16th IEEE International Conference on}. IEEE, 2017.

\bibitem[Kuznetsova et~al.(2020)Kuznetsova, Rom, Alldrin, Uijlings, Krasin, Pont-Tuset, Kamali, Popov, Malloci, Kolesnikov, et~al.]{kuznetsova2020open}
Alina Kuznetsova, Hassan Rom, Neil Alldrin, Jasper Uijlings, Ivan Krasin, Jordi Pont-Tuset, Shahab Kamali, Stefan Popov, Matteo Malloci, Alexander Kolesnikov, et~al.
\newblock The open images dataset v4: Unified image classification, object detection, and visual relationship detection at scale.
\newblock \emph{International journal of computer vision}, 128\penalty0 (7):\penalty0 1956--1981, 2020.

\bibitem[Lang(1995)]{lang1995newsweeder}
Ken Lang.
\newblock {NewsWeeder}: Learning to filter netnews.
\newblock In \emph{Machine Learning Proceedings 1995}, pages 331--339. Morgan Kaufmann, 1995.
\newblock \doi{10.1016/B978-1-55860-377-6.50048-7}.

\bibitem[Lewis et~al.(2004)Lewis, Yang, Rose, and Li]{lewis2004rcv1}
David~D Lewis, Yiming Yang, Tony~G Rose, and Fan Li.
\newblock Rcv1: A new benchmark collection for text categorization research.
\newblock \emph{Journal of machine learning research}, 5\penalty0 (Apr):\penalty0 361--397, 2004.

\bibitem[Lin(2002)]{lin2002divergence}
Jianhua Lin.
\newblock Divergence measures based on the shannon entropy.
\newblock \emph{IEEE Transactions on Information theory}, 37\penalty0 (1):\penalty0 145--151, 2002.

\bibitem[Lin et~al.(2014)Lin, Maire, Belongie, Hays, Perona, Ramanan, Doll{\'a}r, and Zitnick]{lin2014microsoft}
Tsung-Yi Lin, Michael Maire, Serge Belongie, James Hays, Pietro Perona, Deva Ramanan, Piotr Doll{\'a}r, and C~Lawrence Zitnick.
\newblock Microsoft coco: Common objects in context.
\newblock In \emph{European conference on computer vision}, pages 740--755. Springer, 2014.

\bibitem[Liu and Ye(2025)]{liutabpfn}
Siyang Liu and Han-Jia Ye.
\newblock Tabpfn unleashed: A scalable and effective solution to tabular classification problems.
\newblock In \emph{Forty-second International Conference on Machine Learning}, 2025.

\bibitem[Liu et~al.(2019)Liu, Ott, Goyal, Du, Joshi, Chen, Levy, Lewis, Zettlemoyer, and Stoyanov]{liu2019roberta}
Yinhan Liu, Myle Ott, Naman Goyal, Jingfei Du, Mandar Joshi, Danqi Chen, Omer Levy, Mike Lewis, Luke Zettlemoyer, and Veselin Stoyanov.
\newblock Roberta: A robustly optimized bert pretraining approach.
\newblock \emph{arXiv preprint arXiv:1907.11692}, 2019.

\bibitem[Luo et~al.(2025)Luo, Yuan, and Xu]{luo2025time}
Jiaqi Luo, Yuan Yuan, and Shixin Xu.
\newblock Time: Tabpfn-integrated multimodal engine for robust tabular-image learning.
\newblock \emph{arXiv preprint arXiv:2506.00813}, 2025.

\bibitem[Lyubinets et~al.(2018)Lyubinets, Boiko, and Nicholas]{lyubinets2018automated}
Volodymyr Lyubinets, Taras Boiko, and Deon Nicholas.
\newblock Automated labeling of bugs and tickets using attention-based mechanisms in recurrent neural networks.
\newblock In \emph{2018 IEEE Second International Conference on Data Stream Mining \& Processing (DSMP)}, pages 271--275. IEEE, 2018.

\bibitem[Ma et~al.(2024)Ma, Thomas, Hosseinzadeh, Labach, Kamkari, Cresswell, Golestan, Yu, Caterini, and Volkovs]{ma2024tabdpt}
Junwei Ma, Valentin Thomas, Rasa Hosseinzadeh, Alex Labach, Hamidreza Kamkari, Jesse~C Cresswell, Keyvan Golestan, Guangwei Yu, Anthony~L Caterini, and Maksims Volkovs.
\newblock Tabdpt: Scaling tabular foundation models on real data.
\newblock \emph{arXiv preprint arXiv:2410.18164}, 2024.

\bibitem[Maas et~al.(2011)Maas, Daly, Pham, Huang, Ng, and Potts]{maas2011learning}
Andrew Maas, Raymond~E Daly, Peter~T Pham, Dan Huang, Andrew~Y Ng, and Christopher Potts.
\newblock Learning word vectors for sentiment analysis.
\newblock In \emph{Proceedings of the 49th annual meeting of the association for computational linguistics: Human language technologies}, pages 142--150, 2011.

\bibitem[M{\"u}ller et~al.(2021)M{\"u}ller, Hollmann, Arango, Grabocka, and Hutter]{muller2021transformers}
Samuel M{\"u}ller, Noah Hollmann, Sebastian~Pineda Arango, Josif Grabocka, and Frank Hutter.
\newblock Transformers can do bayesian inference.
\newblock \emph{arXiv preprint arXiv:2112.10510}, 2021.

\bibitem[Mur-Labadia et~al.(2026)Mur-Labadia, Muckley, Bar, Assran, Sinha, Rabbat, LeCun, Ballas, and Bardes]{mur2026v}
Lorenzo Mur-Labadia, Matthew Muckley, Amir Bar, Mido Assran, Koustuv Sinha, Mike Rabbat, Yann LeCun, Nicolas Ballas, and Adrien Bardes.
\newblock V-jepa 2.1: Unlocking dense features in video self-supervised learning.
\newblock \emph{arXiv preprint arXiv:2603.14482}, 2026.

\bibitem[Pacheco et~al.(2020)Pacheco, Lima, Salomao, Krohling, Biral, De~Angelo, Alves~Jr, Esgario, Simora, Castro, et~al.]{pacheco2020pad}
Andre~GC Pacheco, Gustavo~R Lima, Amanda~S Salomao, Breno Krohling, Igor~P Biral, Gabriel~G De~Angelo, F{\'a}bio~CR Alves~Jr, Jos{\'e}~GM Esgario, Alana~C Simora, Pedro~BC Castro, et~al.
\newblock Pad-ufes-20: A skin lesion dataset composed of patient data and clinical images collected from smartphones.
\newblock \emph{Data in brief}, 32:\penalty0 106221, 2020.

\bibitem[Park et~al.(2024)Park, Zhang, Yu, Beery, and Huang]{park2024visually}
Seulki Park, Youren Zhang, Stella~X Yu, Sara Beery, and Jonathan Huang.
\newblock Visually consistent hierarchical image classification.
\newblock \emph{arXiv preprint arXiv:2406.11608}, 2024.

\bibitem[{PetFinder}(2019)]{petfinder2019}
{PetFinder}.
\newblock Petfinder.my adoption prediction.
\newblock \url{https://www.kaggle.com/c/petfinder-adoption-prediction}, 2019.
\newblock Kaggle Competition Dataset.

\bibitem[phuc(2024)]{depie2024}
phuc.
\newblock Butterfly image classification, 2024.
\newblock URL \url{https://www.kaggle.com/datasets/phucthaiv02/butterfly-image-classification}.
\newblock Accessed May 5, 2026.

\bibitem[Prokhorenkova et~al.(2018)Prokhorenkova, Gusev, Vorobev, Dorogush, and Gulin]{prokhorenkova2018catboost}
Liudmila Prokhorenkova, Gleb Gusev, Aleksandr Vorobev, Anna~Veronika Dorogush, and Andrey Gulin.
\newblock Catboost: unbiased boosting with categorical features.
\newblock \emph{Advances in neural information processing systems}, 31, 2018.

\bibitem[Qu et~al.(2025)Qu, Holzm{\"u}ller, Varoquaux, and Le~Morvan]{qu2025tabicl}
Jingang Qu, David Holzm{\"u}ller, Ga{\"e}l Varoquaux, and Marine Le~Morvan.
\newblock Tab{ICL}: {A} tabular foundation model for in-context learning on large data.
\newblock In \emph{International Conference on Machine Learning}, 2025.

\bibitem[Qu et~al.(2026)Qu, Holzm{\"u}ller, Varoquaux, and Le~Morvan]{qu2026tabiclv2}
Jingang Qu, David Holzm{\"u}ller, Ga{\"e}l Varoquaux, and Marine Le~Morvan.
\newblock {TabICLv2}: {A} better, faster, scalable, and open tabular foundation model.
\newblock \emph{arXiv preprint arXiv:2602.11139}, 2026.

\bibitem[Reimers and Gurevych(2019{\natexlab{a}})]{reimers-2019-sentence-bert}
Nils Reimers and Iryna Gurevych.
\newblock Sentence-bert: Sentence embeddings using siamese bert-networks.
\newblock In \emph{Proceedings of the 2019 Conference on Empirical Methods in Natural Language Processing}. Association for Computational Linguistics, 11 2019{\natexlab{a}}.
\newblock URL \url{http://arxiv.org/abs/1908.10084}.

\bibitem[Reimers and Gurevych(2019{\natexlab{b}})]{reimers2019sentence}
Nils Reimers and Iryna Gurevych.
\newblock Sentence-bert: Sentence embeddings using siamese bert-networks.
\newblock In \emph{Proceedings of the 2019 conference on empirical methods in natural language processing and the 9th international joint conference on natural language processing (EMNLP-IJCNLP)}, pages 3982--3992, 2019{\natexlab{b}}.

\bibitem[Saleh and Elgammal(2015)]{saleh2015wikiart}
Babak Saleh and Ahmed Elgammal.
\newblock Large-scale classification of fine-art paintings: Learning the right metric on the right feature.
\newblock \emph{arXiv preprint arXiv:1505.00855}, 2015.

\bibitem[Shi et~al.(2021)Shi, Mueller, Erickson, Li, and Smola]{shi2021benchmarking}
Xingjian Shi, Jonas Mueller, Nick Erickson, Mu~Li, and Alexander~J. Smola.
\newblock Benchmarking multimodal {AutoML} for tabular data with text fields.
\newblock In \emph{Proceedings of the Neural Information Processing Systems Track on Datasets and Benchmarks}, 2021.

\bibitem[Shih et~al.(2019)Shih, Wu, Halabi, Kohli, Prevedello, Cook, Sharma, Amorosa, Arteaga, Galperin-Aizenberg, et~al.]{shih2019augmenting}
George Shih, Carol~C Wu, Safwan~S Halabi, Marc~D Kohli, Luciano~M Prevedello, Tessa~S Cook, Arjun Sharma, Judith~K Amorosa, Veronica Arteaga, Maya Galperin-Aizenberg, et~al.
\newblock Augmenting the national institutes of health chest radiograph dataset with expert annotations of possible pneumonia.
\newblock \emph{Radiology: Artificial Intelligence}, 1\penalty0 (1):\penalty0 e180041, 2019.

\bibitem[Sim{\'e}oni et~al.(2025)Sim{\'e}oni, Vo, Seitzer, Baldassarre, Oquab, Jose, Khalidov, Szafraniec, Yi, Ramamonjisoa, et~al.]{simeoni2025dinov3}
Oriane Sim{\'e}oni, Huy~V Vo, Maximilian Seitzer, Federico Baldassarre, Maxime Oquab, Cijo Jose, Vasil Khalidov, Marc Szafraniec, Seungeun Yi, Micha{\"e}l Ramamonjisoa, et~al.
\newblock Dinov3.
\newblock \emph{arXiv preprint arXiv:2508.10104}, 2025.

\bibitem[Sparck~Jones(1972)]{sparck1972statistical}
Karen Sparck~Jones.
\newblock A statistical interpretation of term specificity and its application in retrieval.
\newblock \emph{Journal of documentation}, 28\penalty0 (1):\penalty0 11--21, 1972.

\bibitem[Tang et~al.(2024)Tang, Fang, Zhou, Yang, Zhong, Hu, Kirchhoff, and Karypis]{tang2024autogluon}
Zhiqiang Tang, Haoyang Fang, Su~Zhou, Taojiannan Yang, Zihan Zhong, Tony Hu, Katrin Kirchhoff, and George Karypis.
\newblock Autogluon-multimodal (automm): Supercharging multimodal automl with foundation models.
\newblock \emph{arXiv preprint arXiv:2404.16233}, 2024.

\bibitem[Thoutt(2017)]{thoutt2017winereviews}
Zackary Thoutt.
\newblock Wine reviews.
\newblock \url{https://www.kaggle.com/datasets/zynicide/wine-reviews}, 2017.

\bibitem[Venkataramanan et~al.(2025)Venkataramanan, Pariza, Salehi, Knobel, Gidaris, Ramzi, Bursuc, and Asano]{venkataramanan2025franca}
Shashanka Venkataramanan, Valentinos Pariza, Mohammadreza Salehi, Lukas Knobel, Spyros Gidaris, Elias Ramzi, Andrei Bursuc, and Yuki~M. Asano.
\newblock Franca: Nested matryoshka clustering for scalable visual representation learning.
\newblock \emph{arXiv preprint arXiv:2507.14137}, 2025.

\bibitem[Ye et~al.(2025)Ye, Liu, and Chao]{ye2025closerlooktabpfnv2}
Han-Jia Ye, Si-Yang Liu, and Wei-Lun Chao.
\newblock A closer look at tabpfn v2: Understanding its strengths and extending its capabilities, 2025.
\newblock URL \url{https://arxiv.org/abs/2502.17361}.

\bibitem[{Yelp}(2015)]{yelp2015dataset}
{Yelp}.
\newblock Yelp open dataset.
\newblock \url{https://www.yelp.com/dataset}, 2015.
\newblock Yelp Dataset Challenge.

\bibitem[Yin and Shen(2018)]{yin2018dimensionality}
Zi~Yin and Yuanyuan Shen.
\newblock On the dimensionality of word embedding.
\newblock \emph{Advances in neural information processing systems}, 31, 2018.

\bibitem[Zangari et~al.(2024)Zangari, Marcuzzo, Rizzo, Giudice, Albarelli, and Gasparetto]{zangari2024hierarchical}
Alessandro Zangari, Matteo Marcuzzo, Matteo Rizzo, Lorenzo Giudice, Andrea Albarelli, and Andrea Gasparetto.
\newblock Hierarchical text classification and its foundations: A review of current research.
\newblock \emph{Electronics}, 13\penalty0 (7):\penalty0 1199, 2024.

\bibitem[Zhang et~al.(2015)Zhang, Zhao, and LeCun]{zhang2015character}
Xiang Zhang, Junbo Zhao, and Yann LeCun.
\newblock Character-level convolutional networks for text classification.
\newblock \emph{Advances in neural information processing systems}, 28, 2015.

\bibitem[Zhang et~al.(2021)Zhang, Shen, Dong, Wang, and Han]{zhang2021match}
Yu~Zhang, Zhihong Shen, Yuxiao Dong, Kuansan Wang, and Jiawei Han.
\newblock Match: Metadata-aware text classification in a large hierarchy.
\newblock In \emph{Proceedings of the Web Conference 2021}, pages 3246--3257, 2021.

\bibitem[Zhou et~al.(2020)Zhou, Ma, Long, Xu, Ding, Zhang, Xie, and Liu]{zhou2020hierarchy}
Jie Zhou, Chunping Ma, Dingkun Long, Guangwei Xu, Ning Ding, Haoyu Zhang, Pengjun Xie, and Gongshen Liu.
\newblock Hierarchy-aware global model for hierarchical text classification.
\newblock In \emph{Proceedings of the 58th annual meeting of the association for computational linguistics}, pages 1106--1117, 2020.

\end{thebibliography}

\newpage
\appendix

\section{Datasets \& Experimental Details}
\label{sec:datasets}

\subsection{Compute Resources}
\label{sec:compute_resources}

We use a private compute cluster for our experiments. All CoMET experiments were conducted on L40s GPUs on nodes with 64-256GB of RAM. H100s with 256GB RAM were used to compute results with MM-PFN and TabPFNV2.5.

\subsection{ImageNet Subsets}
\label{sec:imagenetsubsets}

In this paper we use subsets of ImageNet for image classifcation tasks. Our subsets along with their corresponding classes wthin each subset are detailed in Table \ref{tab:imagenetsubsets}.

\begin{table}[ht]
\centering
\caption{}
\label{tab:imagenetdatasets}
\begin{tabular}{lrrr}
\toprule
\textbf{Dataset} & \textbf{Train} & \textbf{Val} & \textbf{Classes} \\
\midrule
Terrier & 12{,}608 & 500 & 10 \\
Snakes & 21{,}871 & 850 & 17 \\
Beetle & 10{,}400 & 400 & 8 \\
Feline & 13{,}000 & 500 & 10 \\
Vehicles & 56{,}956 & 2{,}200 & 44 \\
Dogs & 147{,}873 & 5{,}900 & 118 \\
\bottomrule
\end{tabular}
\end{table}

\begin{table}[htbp]
\centering
\caption{ImageNet subsets used in our experiments and their constituent classes.}
\label{tab:imagenetsubsets}
\renewcommand{\arraystretch}{1.25}
\begin{tabularx}{\textwidth}{@{}lX@{}}
\toprule
\textbf{Subset} & \textbf{Class Names} \\
\midrule

Terrier & Staffordshire bullterrier, American Staffordshire terrier, Bedlington terrier, Border terrier, Kerry blue terrier, Irish terrier, Norfolk terrier, Norwich terrier, Yorkshire terrier, wire-haired fox terrier. \\
\addlinespace

Snakes & thunder snake, ringneck snake, hognose snake, green snake, king snake, garter snake, water snake, vine snake, night snake, boa constrictor, rock python, Indian cobra, green mamba, sea snake, horned viper, diamondback, sidewinder. \\
\addlinespace

Beetle & tiger beetle, ladybug, ground beetle, long-horned beetle, leaf beetle, dung beetle, rhinoceros beetle, weevil. \\
\addlinespace

Feline & tabby, tiger cat, Persian cat, Siamese cat, Egyptian cat, cougar, lynx, leopard, snow leopard, jaguar. \\
\addlinespace

Vehicles & ambulance, amphibian, beach wagon, cab, convertible, fire engine, garbage truck, go-kart, golfcart, jeep, aircraft carrier, canoe, catamaran, container ship, fireboat, gondola, lifeboat, liner, pirate, schooner, airliner, airship, ashcan, electric guitar, forklift, jinrikisha, limousine, minivan, Model T, motor scooter, passenger car, racer, space shuttle, speedboat, sports car, steam locomotive, streetcar, tow truck, tractor, trailer truck, tricycle, unicycle, warplane, wing. \\
\addlinespace

Dogs & Chihuahua, Japanese spaniel, Maltese dog, Pekinese, Shih-Tzu, Blenheim spaniel, papillon, toy terrier, Rhodesian ridgeback, Afghan hound, basset, beagle, bloodhound, bluetick, black-and-tan coonhound, Walker hound, English foxhound, redbone, borzoi, Irish wolfhound, Italian greyhound, whippet, Ibizan hound, Norwegian elkhound, otterhound, Saluki, Scottish deerhound, Weimaraner, Staffordshire bullterrier, American Staffordshire terrier, Bedlington terrier, Border terrier, Kerry blue terrier, Irish terrier, Norfolk terrier, Norwich terrier, Yorkshire terrier, wire-haired fox terrier, Lakeland terrier, Sealyham terrier, Airedale, cairn, Australian terrier, Dandie Dinmont, Boston bull, miniature schnauzer, giant schnauzer, standard schnauzer, Scotch terrier, Tibetan terrier, silky terrier, soft-coated wheaten terrier, West Highland white terrier, Lhasa, flat-coated retriever, curly-coated retriever, golden retriever, Labrador retriever, Chesapeake Bay retriever, German short-haired pointer, vizsla, English setter, Irish setter, Gordon setter, Brittany spaniel, clumber, English springer, Welsh springer spaniel, cocker spaniel, Sussex spaniel, Irish water spaniel, kuvasz, schipperke, groenendael, malinois, briard, kelpie, komondor, Old English sheepdog, Shetland sheepdog, collie, Border collie, Bouvier des Flandres, Rottweiler, German shepherd, Doberman, miniature pinscher, Greater Swiss Mountain dog, Bernese mountain dog, Appenzeller, EntleBucher, boxer, bull mastiff, Tibetan mastiff, French bulldog, Great Dane, Saint Bernard, Eskimo dog, malamute, Siberian husky, dalmatian, affenpinscher, basenji, pug, Leonberg, Newfoundland, Great Pyrenees, Samoyed, Pomeranian, chow, keeshond, Brabancon griffon, Pembroke, Cardigan, toy poodle, miniature poodle, standard poodle, Mexican hairless. \\

\bottomrule
\end{tabularx}
\end{table}

\subsection{Text-only datasets}
\label{sec:textdatasets}

\begin{itemize}

\item \textbf{IMDB} \cite{maas2011learning} consists of movie reviews represented using ELECTRA mean-pooled embeddings. The task is binary sentiment classification (positive vs.\ negative).

\item \textbf{20 Newsgroups} \cite{lang1995newsweeder} consists of newsgroup posts represented using ELECTRA mean-pooled embeddings. The task is topic classification over 20 classes.

\item \textbf{AG News} \cite{zhang2015character} consists of news article titles and descriptions concatenated and represented using ELECTRA mean-pooled embeddings. The task is news-category classification over 4 classes: World, Sports, Business, and Sci/Tech.

\item \textbf{Yelp} (Yelp Dataset Agreement) \cite{zhang2015character} consists of business reviews represented using ELECTRA mean-pooled embeddings. We sample 30{,}000 training examples and 10{,}000 test examples stratified by star rating. The task is review-rating classification over 5 classes.

\end{itemize}

The train/val splits and sequence lengths are outlined in Table \ref{tab:textdatasets}. All texts are truncated to 512 tokens, which is ELECTRA's maximum sequence length. 

\begin{table}[ht]
\centering
\caption{Text classification datasets with their train/test splits, number of classes, and maximum sequence lengths.}
\label{tab:textdatasets}
\begin{tabular}{lrrrcc}
\toprule
\textbf{Dataset} & \textbf{Train} & \textbf{Test} & \textbf{Classes} & \textbf{Max seq.\ len} & \textbf{Avg. seq.\ len}\\
\midrule
IMDB          &  25{,}000 &  25{,}000 &  2 & 512 & 273\\
20 NewsGroups &  11{,}314 &   7{,}532 & 20 & 512 & 175\\
AG News       & 120{,}000 &   7{,}600 &  4 & 256 & 53\\
Yelp          &  30{,}000 &  10{,}000 &  5 & 512 & 129\\
\bottomrule
\end{tabular}
\end{table}

\subsection{Butterflies and RSNA Pneumonia}

The Radiological Society of North America (RSNA) Pneumonia Detection dataset~\citep{shih2019augmenting} consists of chest X-ray images annotated for the presence of pneumonia. It consists of 21,348 training images (16,572/4,776 positive/negative) and 5,336 test (4,100/1,236 positive/negative).

In contrast, the Butterfly Image Classification dataset (CC0 1.0) from Kaggle~\citep{depie2024} contains high-resolution images of various butterfly species across multiple classes, designed for fine-grained visual categorization. We split into 5,200 train and 1,299 test, and the exact label distribution can be seen in Table \ref{tab:butterfly_class}.

\begin{table}[H]
\centering
\small
\setlength{\tabcolsep}{4pt}
\caption{Object classes and their respective image counts in the train and test sets.}
\label{tab:butterfly_class}
\begin{tabular}{lrr @{\hspace{12pt}} lrr}
\hline
\textbf{Class} & \textbf{Train} & \textbf{Test} &
\textbf{Class} & \textbf{Train} & \textbf{Test} \\
\hline
Adonis & 73 & 15 & African Giant Swallowtail & 56 & 19 \\
American Snoot & 62 & 12 & An 88 & 75 & 10 \\
Appollo & 70 & 20 & Atala & 83 & 17 \\
Banded Orange Heliconian & 78 & 19 & Banded Peacock & 68 & 15 \\
Beckers White & 63 & 18 & Black Hairstreak & 67 & 18 \\
Blue Morpho & 59 & 16 & Blue Spotted Crow & 74 & 12 \\
Brown Siproeta & 73 & 26 & Cabbage White & 72 & 18 \\
Cairns Birdwing & 68 & 15 & Checquered Skipper & 75 & 20 \\
Chestnut & 65 & 20 & Cleopatra & 76 & 17 \\
Clodius Parnassian & 70 & 17 & Clouded Sulphur & 69 & 23 \\
Common Banded Awl & 67 & 20 & Common Wood-Nymph & 76 & 14 \\
Copper Tail & 76 & 18 & Crecent & 80 & 17 \\
Crimson Patch & 63 & 9 & Danaid Eggfly & 72 & 22 \\
Eastern Coma & 72 & 21 & Eastern Dapple White & 71 & 21 \\
Eastern Pine Elfin & 70 & 25 & Elbowed Pierrot & 66 & 16 \\
Gold Banded & 60 & 13 & Great Eggfly & 62 & 16 \\
Great Jay & 77 & 17 & Green Celled Cattleheart & 68 & 20 \\
Grey Hairstreak & 67 & 19 & Indra Swallow & 64 & 17 \\
Iphiclus Sister & 71 & 24 & Julia & 66 & 15 \\
Large Marble & 66 & 15 & Malachite & 61 & 12 \\
Mangrove Skipper & 70 & 17 & Mestra & 73 & 13 \\
Metalmark & 64 & 12 & Milberts Tortoiseshell & 80 & 16 \\
Monarch & 66 & 24 & Mourning Cloak & 96 & 35 \\
Orange Oakleaf & 70 & 17 & Orange Tip & 76 & 20 \\
Orchard Swallow & 59 & 17 & Painted Lady & 69 & 9 \\
Paper Kite & 74 & 16 & Peacock & 68 & 16 \\
Pine White & 67 & 19 & Pipevine Swallow & 58 & 26 \\
Popinjay & 66 & 19 & Purple Hairstreak & 60 & 19 \\
Purplish Copper & 74 & 18 & Question Mark & 51 & 26 \\
Red Admiral & 68 & 14 & Red Cracker & 69 & 27 \\
Red Postman & 68 & 21 & Red Spotted Purple & 68 & 18 \\
Scarce Swallow & 81 & 16 & Silver Spot Skipper & 70 & 13 \\
Sleepy Orange & 85 & 22 & Sootywing & 72 & 18 \\
Southern Dogface & 76 & 11 & Straited Queen & 72 & 15 \\
Tropical Leafwing & 71 & 12 & Two Barred Flasher & 69 & 7 \\
Ulyses & 63 & 21 & Viceroy & 69 & 12 \\
Wood Satyr & 60 & 11 & Yellow Swallow Tail & 61 & 14 \\
Zebra Long Wing & 66 & 10 & & & \\
\hline
\end{tabular}
\end{table}

\subsection{Multimodal datasets}
\label{sec:multimodal_datasets}
\begin{itemize}

\item \textbf{Jigsaw Toxicity} (CC0 1.0) \cite{jigsaw2018toxic} consists of online comments paired with 24 demographic identity attributes and 5 user-engagement signals as tabular features, together with ELECTRA embeddings of the comment text. The task is binary toxicity classification.

\item \textbf{Wine Reviews} (CC BY-NC-SA 4.0) \cite{thoutt2017winereviews} consists of wine descriptions paired with structured metadata --- country, province, points, and price --- as tabular features, together with ELECTRA embeddings of the review text. The task is variety classification over 20 classes.

\item \textbf{WikiArt} \cite{saleh2015wikiart} consists of artwork images represented using DINOv3 embeddings, with no accompanying tabular or text features. The task is artistic-style classification over 27 classes.

\item \textbf{PetFinder} \cite{petfinder2019} consists of pet adoption profiles with 12 structured tabular features, DINOv3 image embeddings, and ELECTRA embeddings of the pet description. The task is adoption-speed prediction over 5 classes.

\item \textbf{MM-IMDb} \cite{arevalo2017gated} consists of movie entries with DINOv3 poster-image embeddings and ELECTRA plot-synopsis embeddings, with no tabular features. The task is genre classification over 23 classes.

\item \textbf{Airbnb} (CC BY-NC-SA 4.0) \cite{shi2021benchmarking} consists of Melbourne Airbnb listings with 23 categorical and 23 numerical tabular features, together with ELECTRA embeddings of the listing name, summary, and description. Following TTT \cite{bonnier2024revisiting}, the nightly price is discretized into ten quantile-based classes.

\item \textbf{PAD-UFES-20} (CC BY 4.0) \cite{pacheco2020pad} consists of smartphone-captured skin-lesion images represented using DINOv3 embeddings, paired with 21 clinical tabular features including patient age, lesion diameter, anatomical region, skin type, and medical history. The task is lesion classification over 6 classes.

\item \textbf{Salary India} (CC BY-NC-SA 4.0) \cite{shi2021benchmarking} consists of data scientist job postings with 5 structured tabular features --- company, experience level, location, job designation, and job type --- together with ELECTRA embeddings of the job description and key skills. The task is salary-range classification over 6 classes.

\item \textbf{News Channel} (CC BY-NC-SA 4.0) \cite{shi2021benchmarking} consists of news articles with 16 numeric tabular features capturing NLP statistics and user-engagement signals, paired with ELECTRA embeddings of the article title. The task is channel classification over 6 classes.

\end{itemize}

\subsection{MS Coco \& Open Images Task Construction}
\label{sec:coco_open}

MS COCO~\citep{lin2014microsoft} (CC BY 4.0) and Open Images~\citep{kuznetsova2020open} (CC BY 4.0 / CC BY 2.0) are originally object detection datasets, meaning a single image often has numerous labels associated with it. To construct a single-label dataset, we looked at the most commonly occurring labels, and removed images with co-occurrences. For MS Coco we did this by ignoring the "person" label, due to heavy co-occurrence with other labels, and then taking the 50 most commonly occurring labels. The final dataset consisted of 43 536 training samples and 31 335 test samples. The classes used, and their distribution, can be seen in Table \ref{tab:coco_class}. For Open Images, the process was similar, but since this dataset had a lot more fine-grained labels (e.g., Human arm), the list of ignored common labels (presented in Table \ref{tab:open_exc}) needed to be substantially longer to avoid too many co-occurrences. The final dataset included 600 train and 200 test samples of each class, shown in Table \ref{tab:open_inc} (30 000 /10 000 total).

\begin{table}[h]
\centering
\small
\setlength{\tabcolsep}{4pt}

\caption{Object classes and their respective image counts in the train and test sets.}
\label{tab:coco_class}

\begin{tabular}{lrr @{\hspace{12pt}} lrr @{\hspace{12pt}} lrr}
\hline
\textbf{Class} & \textbf{Train} & \textbf{Test} & 
\textbf{Class} & \textbf{Train} & \textbf{Test} & 
\textbf{Class} & \textbf{Train} & \textbf{Test} \\
\hline
surfboard & 2078 & 693 & giraffe & 1639 & 546 & skis & 1636 & 546 \\
train & 1469 & 490 & skateboard & 1462 & 487 & airplane & 1410 & 470 \\
bird & 1334 & 445 & horse & 1324 & 441 & clock & 1291 & 430 \\
car & 1054 & 352 & boat & 997 & 332 & tie & 964 & 322 \\
toilet & 921 & 307 & dog & 915 & 305 & bench & 799 & 266 \\
motorcycle & 790 & 263 & tennis racket & 735 & 245 & cell phone & 706 & 235 \\
bed & 626 & 208 & cat & 622 & 208 & truck & 598 & 200 \\
bus & 508 & 169 & umbrella & 505 & 168 & traffic light & 481 & 160 \\
pizza & 466 & 155 & bowl & 448 & 149 & dining table & 440 & 146 \\
vase & 427 & 142 & sports ball & 351 & 117 & sink & 350 & 116 \\
backpack & 317 & 106 & handbag & 311 & 104 & bottle & 300 & 100 \\
baseball bat & 284 & 94 & chair & 282 & 94 & oven & 268 & 89 \\
baseball glove & 262 & 87 & cake & 256 & 85 & remote & 220 & 73 \\
cup & 219 & 73 & bicycle & 214 & 71 & laptop & 184 & 62 \\
tv & 169 & 56 & knife & 160 & 54 & book & 142 & 47 \\
potted plant & 105 & 35 & couch & 98 & 32 & fork & 83 & 28 \\
wine glass & 58 & 20 & spoon & 57 & 19 & & & \\
\hline
\end{tabular}
\end{table}

\begin{table}[ht]
\centering
\footnotesize

\begin{subtable}[t]{0.48\textwidth}
\centering
\begin{tabular}{lll}
\hline
Beer & Bench & Bicycle \\
Bird & Boat & Book \\
Bottle & Bus & Butterfly \\
Cake & Camera & Car \\
Castle & Cat & Chair \\
Coffee cup & Computer monitor & Dog \\
Drum & Duck & Fish \\
Fixed-wing aircraft & Flag & Flowerpot \\
Football & Guitar & Helmet \\
Horse & Houseplant & Laptop \\
Microphone & Mobile phone & Motorcycle \\
Palm tree & Picture frame & Poster \\
Rose & Salad & Sculpture \\
Shelf & Stairs & Street light \\
Swimming pool & Table & Tent \\
Toy & Train & Truck \\
Van & Wine glass & \\
\hline
\end{tabular}
\caption{Included Classes (50)}
\label{tab:open_inc}
\end{subtable}
\hfill
\begin{subtable}[t]{0.48\textwidth}
\centering
\begin{tabular}{p{0.95\linewidth}}
\hline
Man, Woman, Girl, Boy, Person; Human face, Human body, Human arm, Human hair, Human head, Human leg, Human mouth, Human eye, Human nose, Human hand, Human foot, Human ear, Human beard; Clothing, Footwear, Suit, Dress, Jeans, Hat, Shirt, Shorts, Glove, Coat, Sock, Tie, Swimwear, Trousers, Scarf, Fashion accessory; Sunglasses, Glasses, Mammal, Land vehicle; Vehicle, Furniture, Animal, Plant, Tree, Flower; Food, Drink, Building, House, Tower, Skyscraper; Wheel, Tire; Sports equipment, Auto part, Tableware, Musical instrument, Insect; Bicycle wheel, Vehicle registration plate; Jacket, Sports uniform, Goggles; Window, Door; Snack, Fast food, Baked goods, Dessert, Fruit, Vegetable; Bicycle helmet, Wine, Sun hat, Dairy Product, Ball (Object); Bookcase, Desk, Office building \\
\hline
\end{tabular}
\caption{Ignored common labels (66)}
\label{tab:open_exc}
\end{subtable}

\caption{Open Images class split}
\label{tab:open_split}

\end{table}

\subsection{Amazon MM and iNaturalist dataset Preprocessing}
\label{sec:preprocess_hier}

The iNaturalist dataset contains 2.8M images. The hierarchy is severely unbalanced, so in order to avoid having the accuracy metrics be dominated by how well TabICL predicts the most popular branches we downsample it to 796,497 training images spanning 2,896 classes. To obtain a balanced hierarchy, we enforce that within each node, no child has more than a 4× smaller sample count than any other child. This is achieved through a bottom-up pruning procedure starting from the leaves, where low-sample nodes are iteratively pruned from the tree until the criterion is satisfied. The resulting tree is relatively well-balanced and suitable for benchmarking TabICL.
We use six tabular features per image: \textit{latitude, longitude, month\_sin, month\_cos, year, and location\_uncertainty}.

The Amazon MM dataset, derived from Amazon Reviews~\citep{hou2024bridging}, includes the features listed in Table \ref{tab:amazon_features}. Text features were extracted using Sentence-BERT on all texts concatenated. Image features were extracted using DINOv3. 9 top-level categories were used. These were: \textit{Arts crafts and sewing}, \textit{Automotive}, \textit{Beauty and personal care}, \textit{Cell phones and accessories}, \textit{Clothing shoes and jewelry}, \textit{Electronics}, \textit{Home and kitchen}, \textit{Sports and outdoors}, \textit{Tools and home improvement}. We then take the following steps to clean and balance the dataset: We limit the depth to 4, and refer to the different levels as L0 (the 9 categories described above), L1, L2, and L3. We then drop L1 classes with fewer than 5000 items, L2 classes with fewer than 500, and L3 with fewer than 100. Additionally, we drop all L1 nodes that don't have any children, meaning the leaf-nodes are strictly located in L2 or L3. Within each parent node (subtask), we allow for a max class imbalance ratio of 5, subsampling larger classes until they are at most 5x the minority class. 

\begin{table}[H]
\centering
\caption{Product features of the Amazon MM dataset.}
\label{tab:amazon_features}
\begin{tabular}{ll}
\toprule
\textbf{Name} & \textbf{Type} \\
\midrule
Title       & Text \\
Description & Text \\
Features    & Text \\
Product Image & Image \\
Price       & Numeric \\
Rating      & Numeric \\
Store       & Categorical \\
\bottomrule
\end{tabular}
\end{table}

\newpage 
\section{Discussion on TabPFN}
\label{sec:tabpfn_discussion}

\subsection{MM-TabPFN's class size limit}
\label{sec:mmtabicl_discussion}

MM-PFN requires backpropogation through TabPFN to finetune its modality specific encoders. In its current form, TabPFN is not designed to perform backprop with more than 10 classes. There exists an extension to TabPFN called many-class-classifier that uses multiple evaluations of TabPFN at inference time, but not training, \href{https://github.com/PriorLabs/tabpfn-extensions/blob/main/examples/many_class/many_class_classifier_example.py}{https://github.com/PriorLabs/tabpfn-extensions/blob/main/examples/many\_class/many\_class\_classifier\_example.py}. Hence, we cannot evaluate the MM-PFN paper on large scale datasets with more than 10 classes.  

\subsection{TabPFNv2.5 vs TabICLv2 on Unimodal datasets}
\label{sec:tabpfnvstabicl_perf}

\begin{figure}[H]
    \centering
    \includegraphics[width=\linewidth]{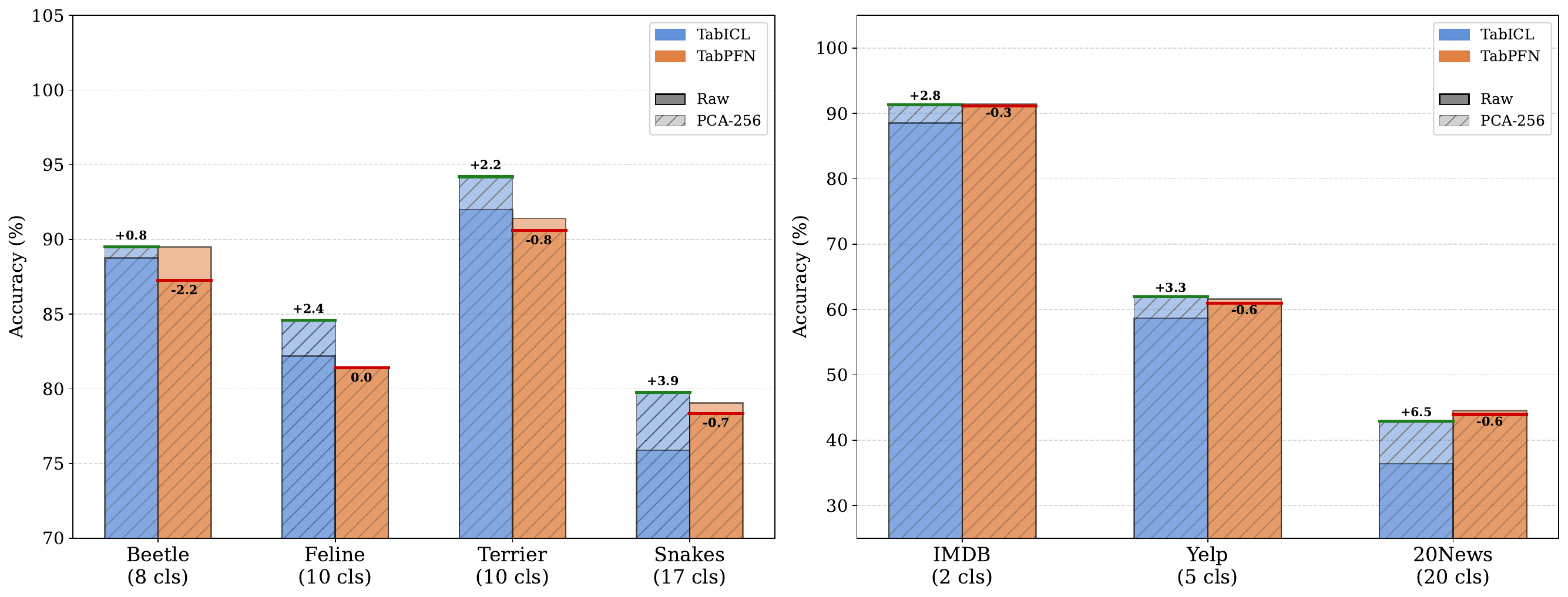}
    \caption{Performance of TabICL vs TabPFN on a few single modality datasets.}
    \label{fig:tabpfnvstabicl_accuracy}
\end{figure}

In Figure \ref{fig:tabpfnvstabicl_accuracy}, we find that TabPFNv2.5 performs slightly worse than TabICLv2 with PCA on image datasets, but matches TabICLv2 on text datasets. We also see that TabPFNv2.5 does not benefit from dimensionality reduction, compared to TabICLv2, which receives a significant boost in performance.

\subsection{TabPFNv2.5 vs TabICLv2 inference times}
\label{sec:tabpfnvstabicl_perf}

In Table \ref{tab:inference-time}, we compare inference times of TabICLv2 against TabPFNv2.5 on both raw \texttt{CLS} tokens and PCA versions. We find that TabICLv2 is significantly faster, showing that it's a much better choice for large-scale multimodal datasets.

\begin{table}[ht]
\caption{Inference time for TabICL vs.\ TabPFN on a few single modality datasets. Lower is better.}
\centering
\label{tab:inference-time}
\begin{tabular}{lcccc}
\toprule
\textbf{Dataset} & \textbf{TabICL+\texttt{CLS}} & \textbf{TabICL+PCA} & \textbf{TabPFN+\texttt{CLS}} & \textbf{TabPFN+PCA} \\
\midrule
\multicolumn{5}{l}{\textit{ImageNet Subsets}} \\
Beetle (8 classes)   & \textbf{18 s}  & \textbf{6 s}   & 41 s   & 14 s   \\
Feline (10 classes)  & \textbf{24 s}  & \textbf{8 s}   & 62 s   & 22 s   \\
Terrier (10 classes) & \textbf{23 s}  & \textbf{8 s}   & 60 s   & 21 s   \\
Snakes (17 classes)  & \textbf{1.5 m} & \textbf{34 s}  & 95 m   & 35 m   \\
\midrule
\multicolumn{5}{l}{\textit{Text}} \\
IMDB (2 classes)     & \textbf{31 m}  & \textbf{12 m}  & 2.9 h  & 52 m   \\
Yelp (5 classes)     & \textbf{15 m}  & \textbf{6 m}   & 1.3 h  & 28 m   \\
20News (20 classes)  & \textbf{7 m}   & \textbf{2 m}   & 5.4 h  & 2.0 h  \\
\bottomrule
\end{tabular}
\end{table}

\newpage
\section{Comparing foundation model embeddings to TabICL's training data}
\label{sec:comparison}

\begin{figure}[H]
    \centering
    \includegraphics[width=0.80\linewidth]{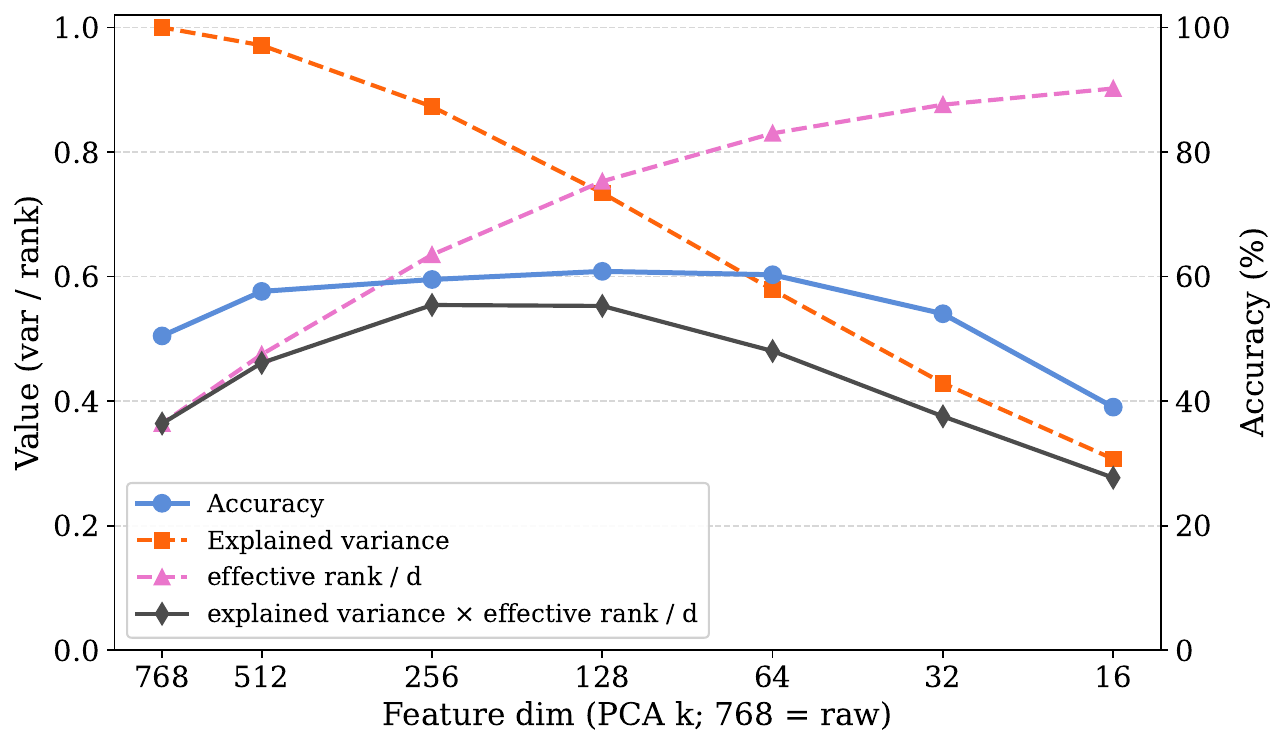}
    \caption{Explained variance, normalized effective rank and their product on subsampled iNaturalist across PCA dims.}
    \label{fig:pcasweep}
\end{figure}

We aim to study why PCA helps TabICL perform better on embeddings from frozen backbones. To do this, we create a stratified-by-class 100K subset of the iNaturalist dataset defined in Section~\ref{sec:preprocess_hier} and evaluate performance across a range of PCA dimensions, alongside effective rank and explained variance, shown in Figure~\ref{fig:pcasweep}. We only use DinoV3 features for this experiment, no tabular data.

We observe accuracy gains from PCA up to 64-dimensional embeddings, after which performance degrades as the latent becomes over-compressed at 32 and 16 dimensions. The curve plotting the product of explained variance and effective rank closely mirrors the accuracy curve, suggesting a tradeoff between preserving information in the latents and accommodating TabICL's preference for high-effective-rank embeddings.

We hypothesize that PCA helps because of a distribution shift between the synthetic data TabICLv2 was pretrained on and the embeddings produced by modality-specific backbones. To test this, we sample 20{,}000 datasets that approximate TabICL's pretraining distribution and compare them to the iNaturalist embeddings actually fed into the model.

The TabICLv2 paper introduces a richer prior than TabICLv1, combining eight families of random functions with a new graph-sampling mechanism. However, the implementation of this v2 prior has not yet been publicly released. Only the v1 generators, which contain a mixture of MLP-based and tree-ensemble structural causal models are available. We therefore sample from the v1 generators while matching the curriculum schedule described in the v2 paper: dataset sizes are drawn across three stages (1{,}024 samples fixed; 400--10{,}240 log-uniform; and 400--60{,}000 log-uniform) in proportion to the published v2 training-step counts ($\approx 91\% / 7\% / 2\%$). This provides the closest publicly available approximation to the distribution TabICLv2 encounters during pretraining.

\begin{figure}[H]
    \centering
    \includegraphics[width=0.80\linewidth]
    {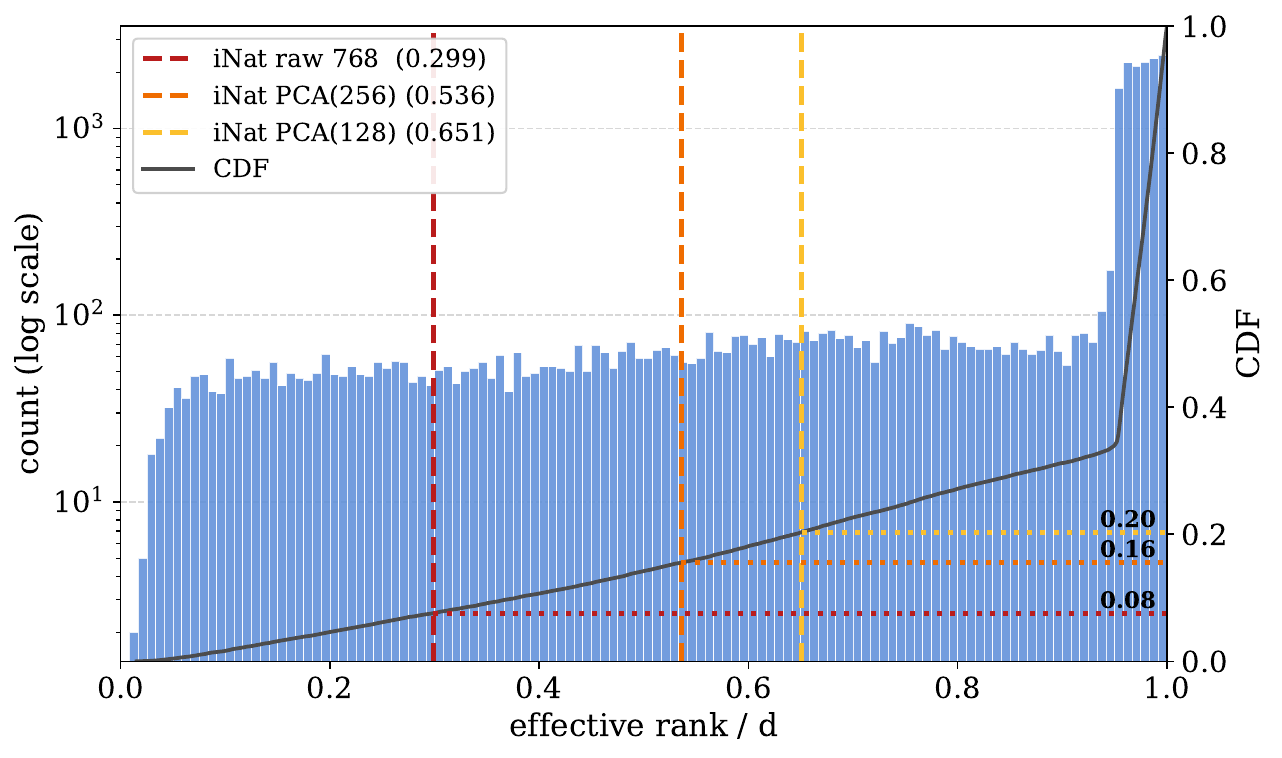}
    \caption{Comparing normalized effective rank between samples from TabICL's prior to iNaturalist DinoV3 CLS tokens with various PCA settings. The log CDF of the histogram is shown in black, highlighting the distribution of the TabICL dataset's effective ranks}
    \label{fig:effrank}
\end{figure}

Figure~\ref{fig:effrank} compares iNaturalist embeddings against these synthetic samples. The embeddings seen during TabICLv2 pretraining are mostly isotropic, with more than 60\% of samples exhibiting effective-ranks above 0.90. In contrast, raw DINOv3 CLS embeddings lie in the left tail of the prior's distribution, having lower effective rank. Applying PCA gradually shifts the DINOv3 embeddings toward the bulk of the prior distribution. This suggests that the gains from PCA arise because TabICL is biased toward high-effective-rank embeddings, reflecting the statistical structure of its synthetic pretraining distribution.

\section{Additional Results}
\label{sec:add_res}

\subsection{Impact of PCA dimensionality}
\label{sec:pca_dim_choice}

\begin{figure}[H]
    \centering
    \includegraphics[width=\linewidth]{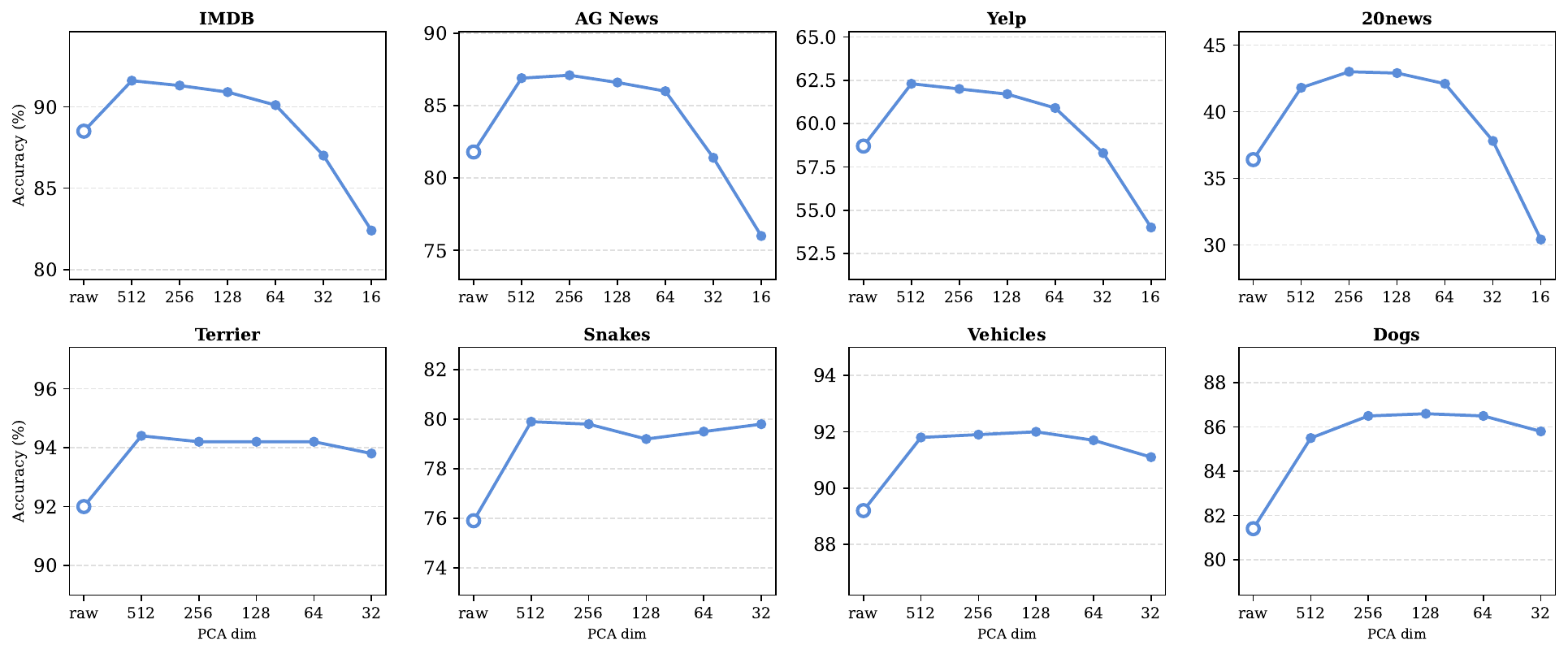}
    \caption{TabICL with varying PCA settings on a image and text only datasets. We find that even applying a slight amount of PCA results in a significant performance boost. PCA dimensions of 512, 256 and 128 are viable. Hence, in this paper, we either use PCA dims 256 or 128 in all experiments.}
    \label{fig:all_pca_sweep}
\end{figure}

\subsection{Testing CoMET with different backbones}

We also run CoMET on Franca \cite{venkataramanan2025franca} for image encoding and RoBERTa \cite{liu2019roberta} for text encoding, shown in \ref{tab:otherbackbones}. Our results on different backbones show that our method is not limited to DinoV3 and ELECTRA, and that the benefits of PCA are due to TabICL, not the backbones.

\begin{table}[h]
\centering
\caption{Results across a few multimodal and single modality datasets for RoBERTa and Franca.}
\label{tab:otherbackbones}
\begin{tabular}{lcc}
\toprule
\textbf{Dataset / Modality} & \textbf{Raw} & \textbf{PCA-256} \\
\midrule
PetFinder (tabular + franca) & 0.3730 & \textbf{0.3991} \\
PetFinder (tabular + roberta) & 0.4057 & \textbf{0.4305} \\
PetFinder (tabular + franca + roberta) & 0.3648 & \textbf{0.4395} \\
\midrule
MM-IMDb (franca) & 0.5487 & \textbf{0.5499} \\
MM-IMDb (roberta) & 0.6538 & \textbf{0.6574} \\
MM-IMDb (franca + roberta) & 0.6495 & \textbf{0.6607} \\
\midrule
Airbnb-TTT (tabular + roberta) & 0.4128 & \textbf{0.4563} \\
\midrule
Snakes (franca) & 0.7212 & \textbf{0.7576} \\
\midrule
Vehicles (franca) & 0.8705 & \textbf{0.8732} \\
\bottomrule
\end{tabular}
\end{table}

\subsection{PALPooling}

\begin{figure}[H]
    \centering
    \includegraphics[width=0.96\linewidth]{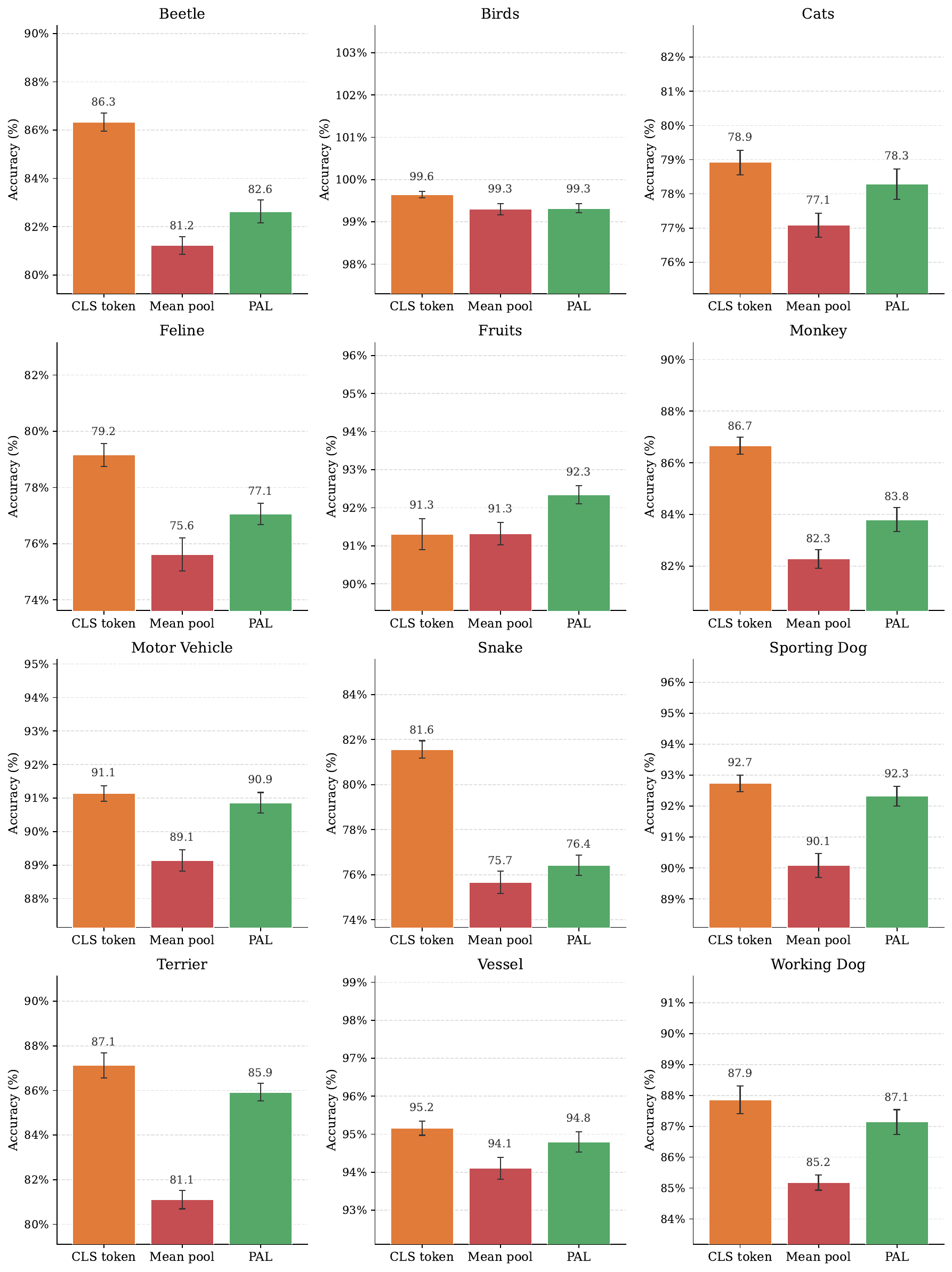}
    \caption{Performance of PALPooling for imagenet subsets where we expect the \texttt{CLS}-token to be strong. Still, we see PALPooling consistently improve over mean-pooling, which is the starting point of the algorithm.}
    \label{fig:pal_pooling_imagenet}
\end{figure}

\begin{figure}[H]
    \centering
    
    \begin{subfigure}{0.99\linewidth}
        \centering
        \includegraphics[width=\linewidth]{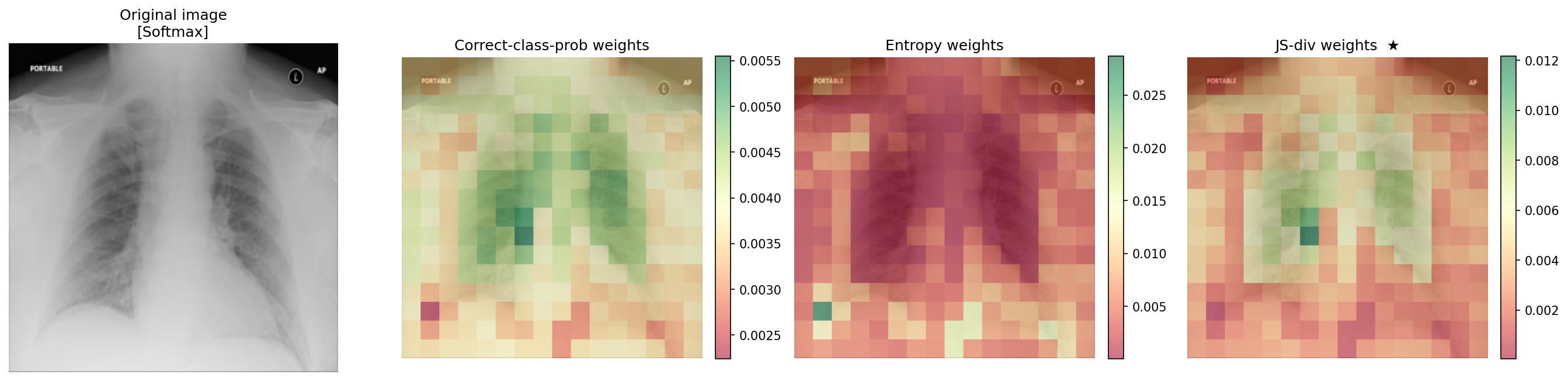}
        \caption{Positive for Pneumonia}
        \label{fig:sub1}
    \end{subfigure}
    
    \vspace{0.5em}
    
    \begin{subfigure}{0.99\linewidth}
        \centering
        \includegraphics[width=\linewidth]{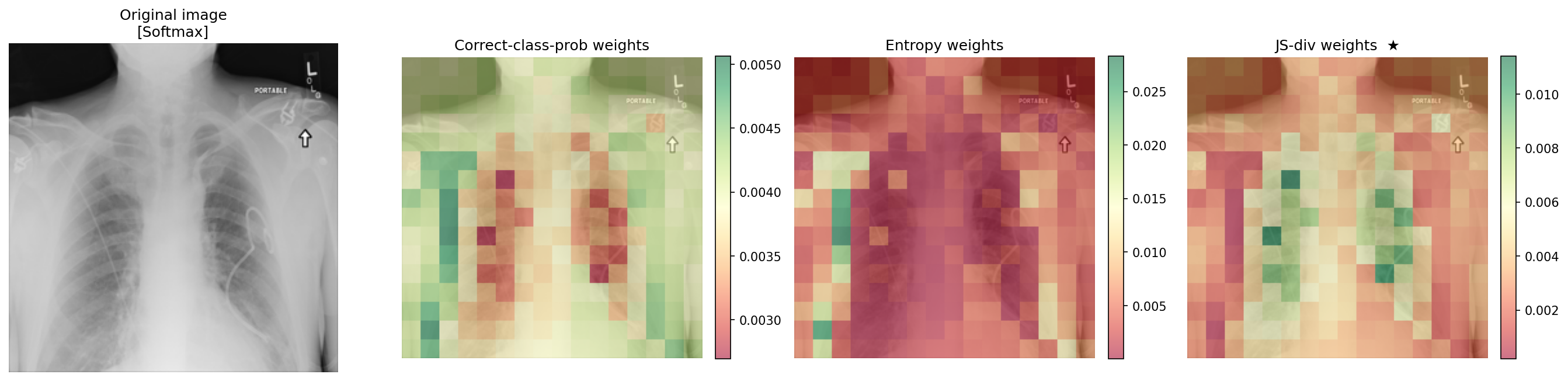}
        \caption{Negative for Pneumonia}
        \label{fig:sub2}
    \end{subfigure}
    
    \caption{A comparison of different methods for generating PALs for RSNA-Pneumonia, where the lungs are the area of interest. While the correct-class probabilities work well for the positive case, it explicitly down-weight the lungs for negative samples. In the third column, we show the results when weights are derived from Shannon's entropy on the predicted label distribution, down-weighting tokens that are closer to 50/50. We see that this approach does not work well for this dataset, likely because the label balance is close to 80/20\% . This means that a piece of background could have a sharper (negative) prediction, compared to a piece of lung tissue. The right-most column shows the weights derived from the JS divergence between the token predictions and the prior label frequency. We see that the lungs are clearly highlighted, indicating that the token predictions are distinct from the class prior.}
    \label{fig:bad_scoring_functions}
\end{figure}

\begin{figure}[H]
    \centering
    \includegraphics[width=0.7\linewidth]{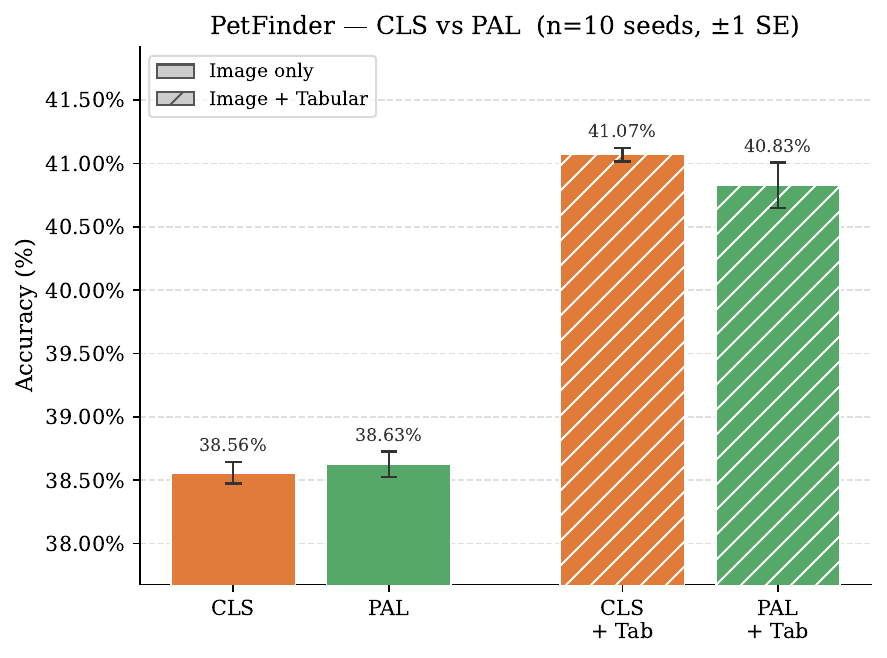}
    \caption{Example of when modality-agnostic pooling can become detrimental. The experiments are averaged over 10 seeds with error bars corresponding to 1 standard error. Experiments used a PCA dimension of 256 and 4 estimators in the TabICL model. Although we observe PALPooling perform similarly or better than the \texttt{CLS} token on the image-only task, we see that it has noticeably worse image + tabular performance.}
    \label{fig:petfinder_cls_pal}
\end{figure}

\begin{figure}[H]
    \centering
    \includegraphics[width=0.9\linewidth]{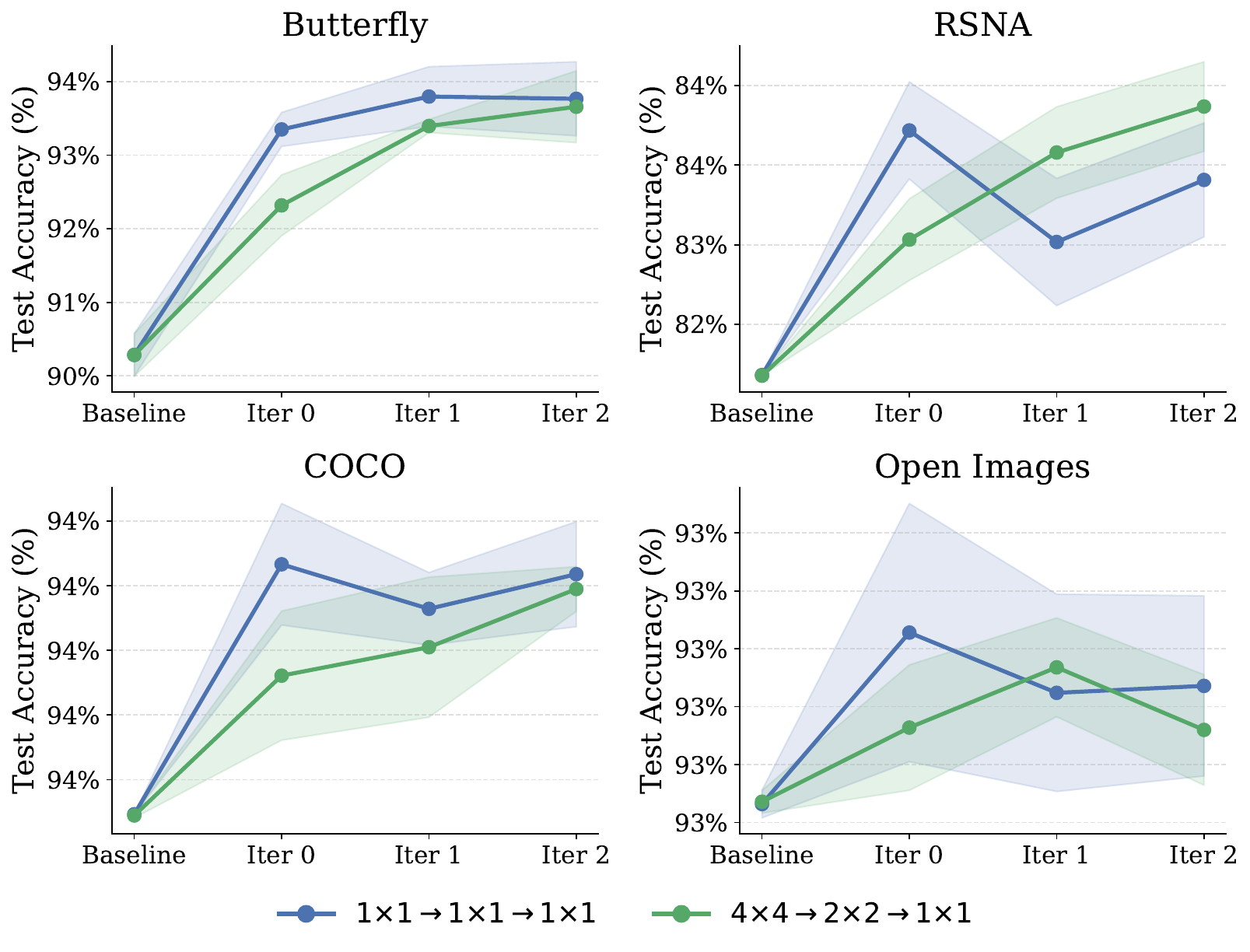}
    \caption{The impact of running PALPooling iteratively across the image datasets, as well as the choice effects of applying coarser pooling.}
    \label{fig:pal_iterations}
\end{figure}

\subsection{Hierarchical classification}

\subsubsection{Timing TabICL on Inaturalist}

In Table \ref{tab:inaturalist}, we compare the inference times of hierarchical and flat TabICL on iNaturalist under different sample caps. We find that training a hierarchical classifier with a cap of 100k–200k samples per TabICL fit is significantly faster than a flat TabICL classifier over all 2,896 classes, even when both use the same overall budget of 100k–200k samples.

This improvement in speed/performance arises because the hierarchy decomposes the task into more manageable sub-problems for TabICL. The speedups stem from the use of mixed-radix ensembling in TabICL to handle more than 10 classes, which forces the flat baseline to perform significantly more forward passes as the number of classes increase. While the hierarchical model still performs forward passes with up to 200k context near the root of the tree, it operates over far fewer classes per TabICL fit than the flat approach, resulting in speed gains.

\begin{table}[h]
\centering
\caption{Inference (predict) time on Naturalist (2{,}896 classes, 28{,}960 val images, mean (std) over 5 seeds, seconds).}
\label{tab:inaturalist}
\begin{tabular}{lcc}
\toprule
\textbf{Model} & \textbf{Acc} $\uparrow$ & \textbf{Predict time (s)} $\downarrow$ \\
\midrule
H-CoMET (cap=100K) & 82.24 (0.11) & 227.1 (1.1) \\
H-CoMET (cap=200K) & 82.50 (0.04) & 315.1 (0.6) \\
F-CoMET ($n$=100K) & 65.08 (0.24) & 704.9 (1.7) \\
F-CoMET ($n$=200K) & 69.54 (0.23) & 1{,}464.4 (2.6) \\
\bottomrule
\end{tabular}
\end{table}

\subsubsection{Text Benchmarks}
\label{sec:text_benchmarks}

\begin{table}[h]
\centering
\caption{Performance of H-CoMET, compared to our baselines and SOTA hierarchical text models.}
\label{tab:hier_text_app}
\begin{tabular}{lcccccc}
\toprule
\multirow{2}{*}{\textbf{Model}} & \multicolumn{2}{c}{\textbf{WOS}} & \multicolumn{2}{c}{\textbf{Amazon}} & \multicolumn{2}{c}{\textbf{Bugs}} \\
\cmidrule(lr){2-3} \cmidrule(lr){4-5} \cmidrule(lr){6-7}
& \textbf{Acc} $\uparrow$ & \textbf{F\textsubscript{1}} $\uparrow$ & \textbf{Acc} $\uparrow$ & \textbf{F\textsubscript{1}} $\uparrow$ & \textbf{Acc} $\uparrow$ & \textbf{F\textsubscript{1}} $\uparrow$ \\
\midrule
MATCH & 59.32 (1.61) & 66.72 (1.45) & 87.17 (0.87) & 90.39 (0.28) & 36.24 (4.91) & 36.35 (5.59) \\
HiAGM & 65.13 (1.59) & 75.44 (0.65) & 87.35 (0.44) & 90.29 (0.30) & 44.82 (0.69) & 52.18 (0.45) \\
HBGL & \textbf{80.14 (0.02)} & \textbf{82.21 (0.04)} & 84.05 (0.02) & 87.96 (0.12) & 57.63 (0.03) & 57.09 (0.32) \\
GACaps & 73.76 (0.78) & 80.63 (0.50) & 86.73 (0.19) & 90.58 (0.10) & 49.16 (0.52) & 54.30 (1.10) \\
BERT & 77.12 (0.67) & 79.81 (0.59) & 89.54 (0.20) & \textbf{92.12} (0.14) & 50.70 (1.13) & 52.04 (0.93) \\
\midrule
H-CoMET & 75.53 (0.38) & 75.13 (0.48) & \textbf{89.85} (0.04) & 91.33 (0.04) & \textbf{76.61} (0.44) & \textbf{78.54} (0.24) \\
F-CoMET & 73.41 (0.28) & 72.05 (0.41) & 87.17 (0.09) & 87.26 (0.09) & 31.44 (0.41) & 23.56 (0.47)\\
F-MLP & 76.16 (0.26) & 75.94 (0.32) & 88.48 (0.07) & 88.57 (0.10) & 31.40 (0.91) & 20.84 (1.94)\\
\bottomrule
\end{tabular}
\end{table}

\begin{figure}[h]
    \centering
    \begin{subfigure}[t]{0.99\textwidth}
        \centering
        \includegraphics[width=\linewidth,height=\linewidth]{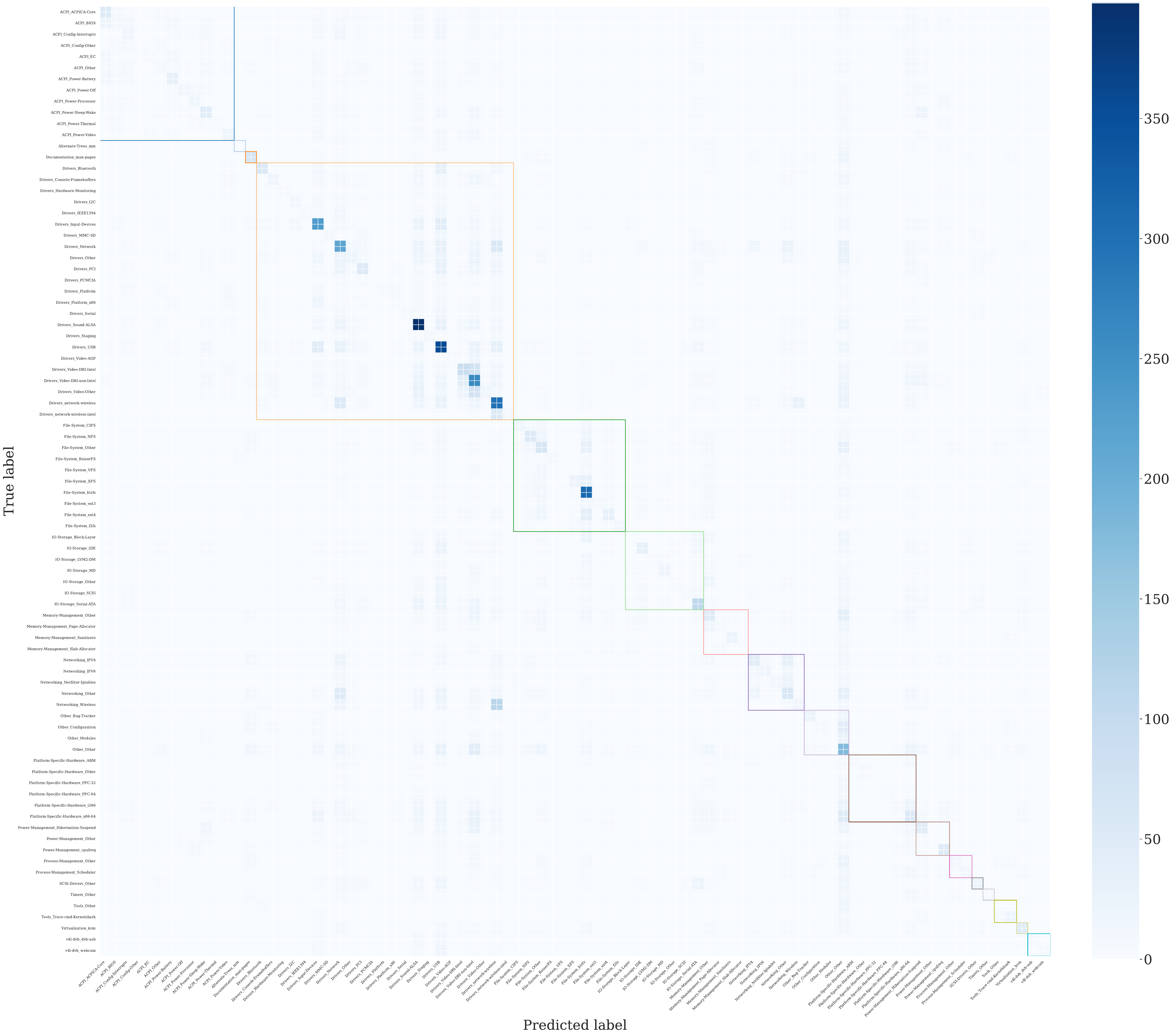}
        \caption{Original}
    \end{subfigure}

    \caption{Overview confusion matrix for TabICL-Flat on the Bugs dataset. Coloured blocks represent subcategories, a selection of which are shown in Figure \ref{fig:cmap_flat_sub}.}
    \label{fig:cmap_flat}
\end{figure}

\begin{figure}[h]
    \centering
    \begin{subfigure}[t]{0.49\textwidth}
        \centering
        \includegraphics[width=\linewidth,height=\linewidth]{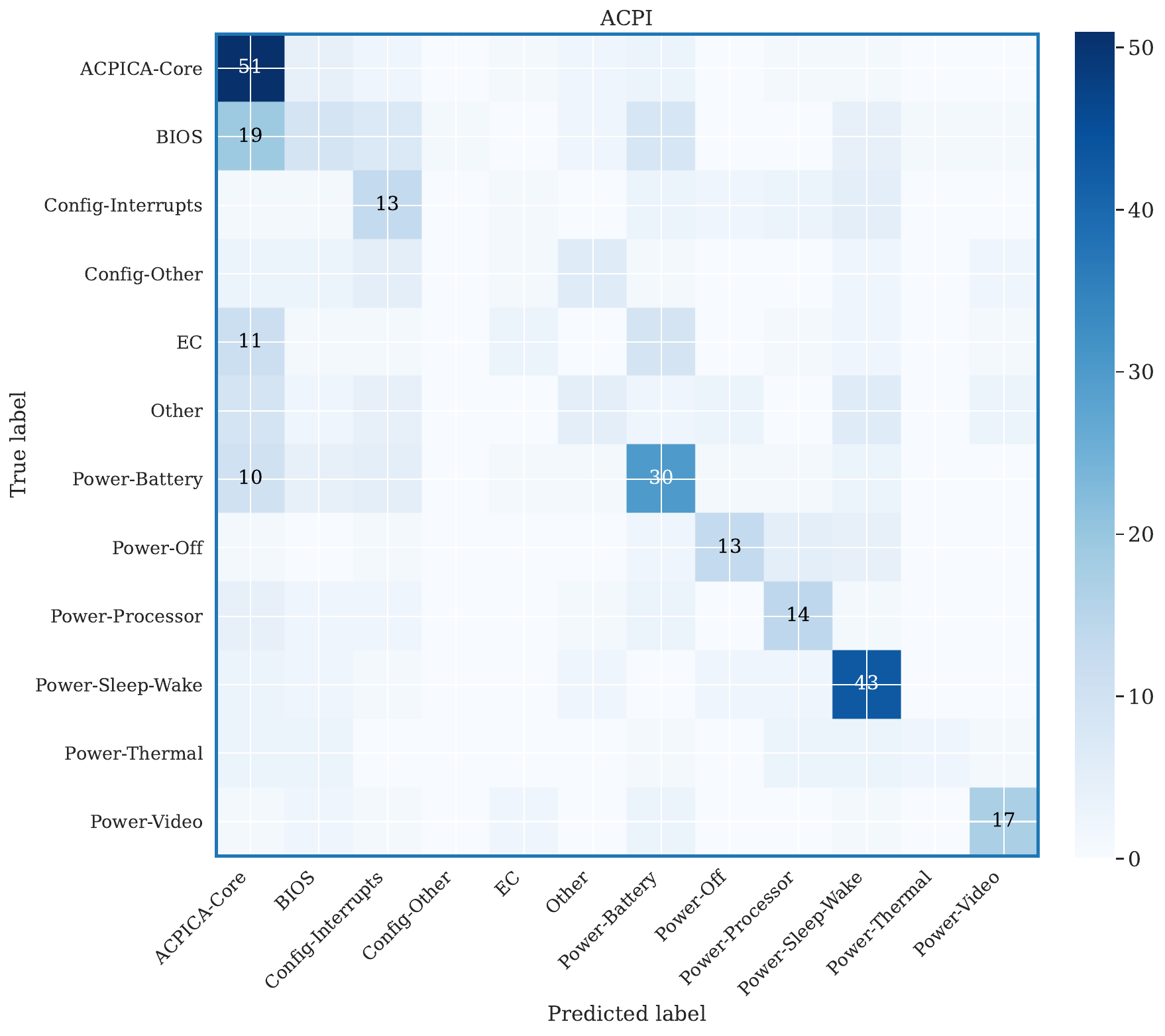}
        \caption{TabICL-Flat ACPI Prediction}
    \end{subfigure}
    \hfill
    \begin{subfigure}[t]{0.49\textwidth}
        \centering
        \includegraphics[width=\linewidth,height=\linewidth]{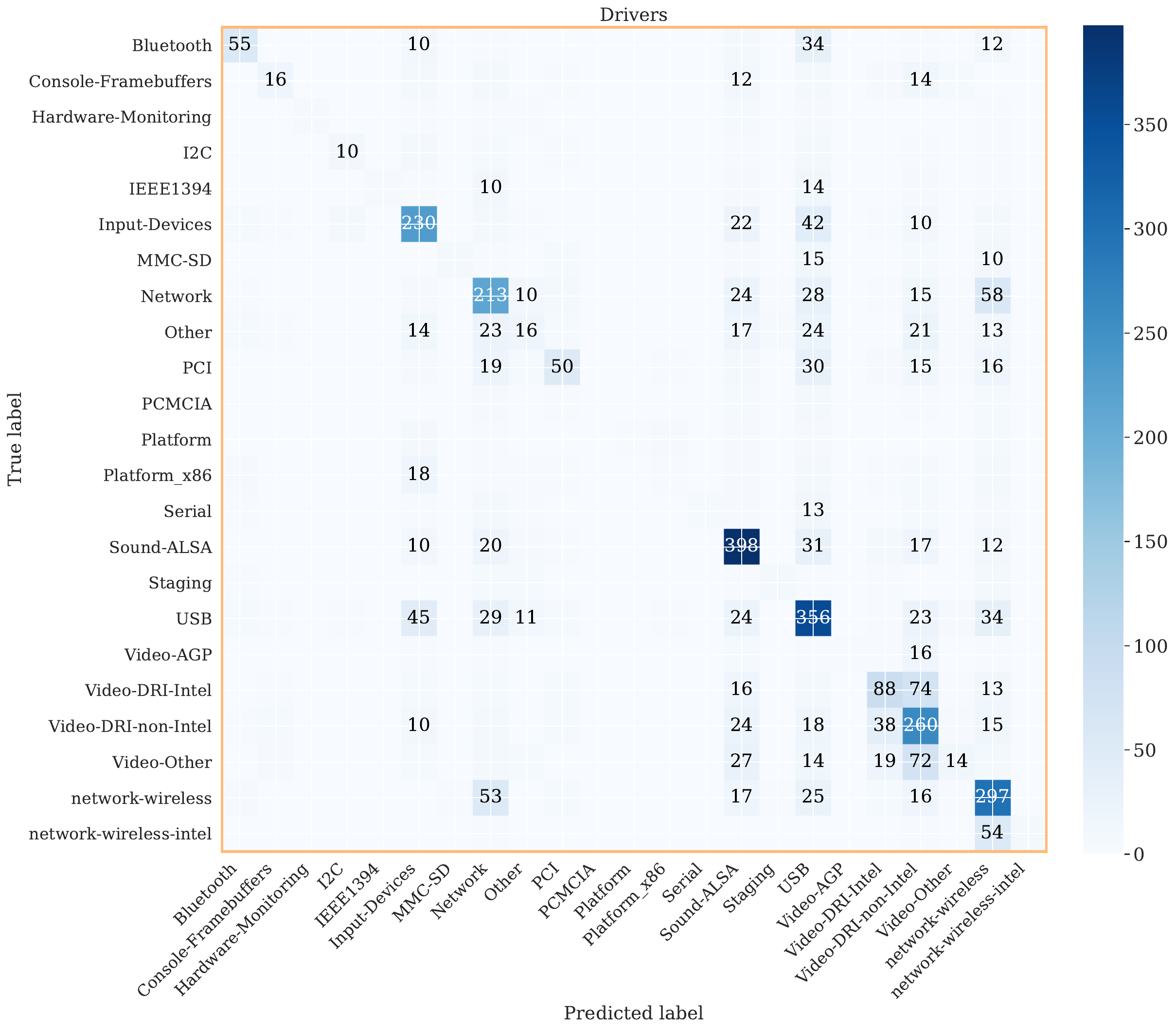}
        \caption{TabICL-Flat Drivers Prediction}
    \end{subfigure}
    \hfill
    \begin{subfigure}[t]{0.49\textwidth}
        \centering
        \includegraphics[width=\linewidth,height=\linewidth]{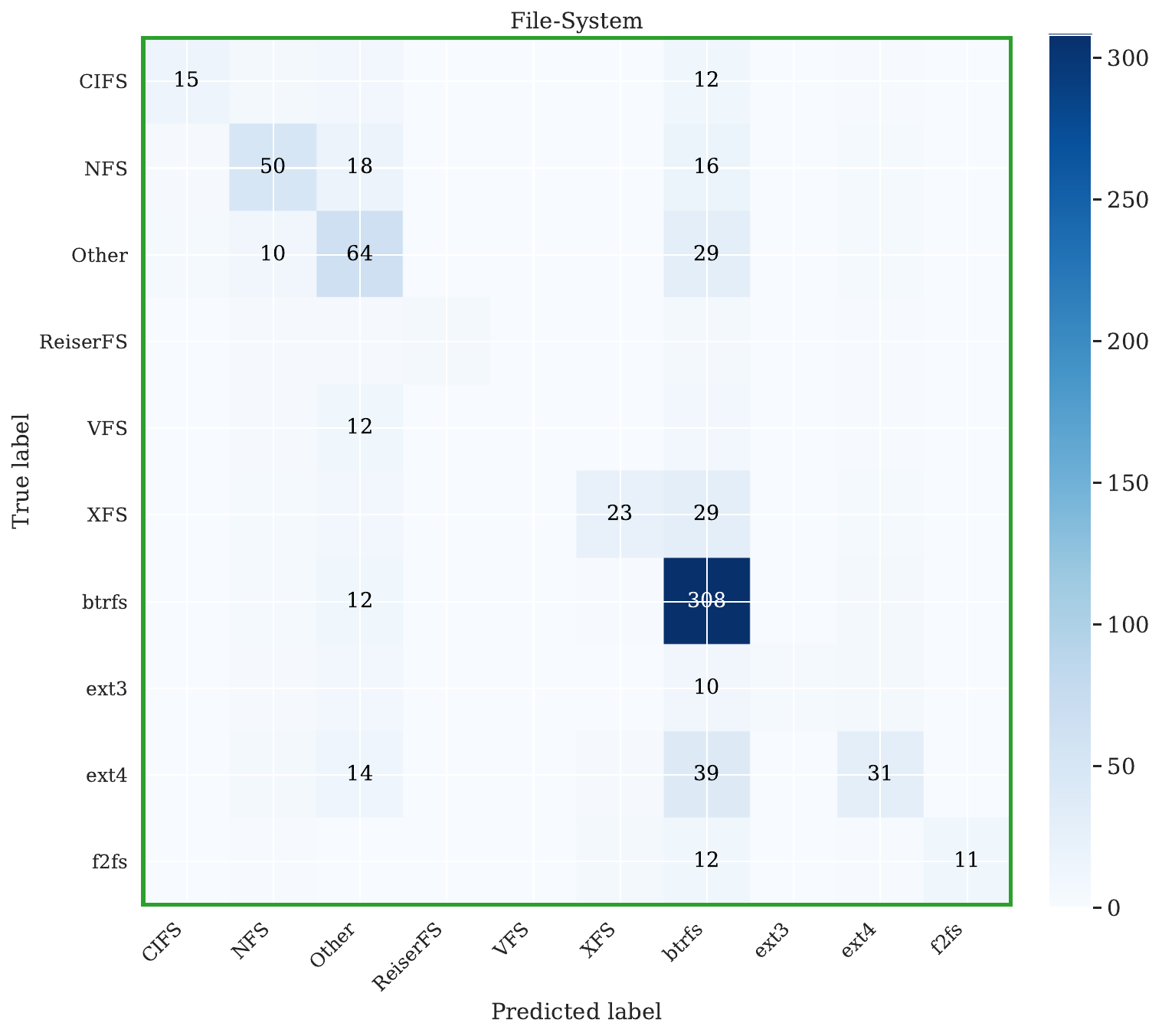}
        \caption{TabICL-Flat File-System Predictions}
    \end{subfigure}
    \hfill
    \begin{subfigure}[t]{0.49\textwidth}
        \centering
        \includegraphics[width=\linewidth,height=\linewidth]{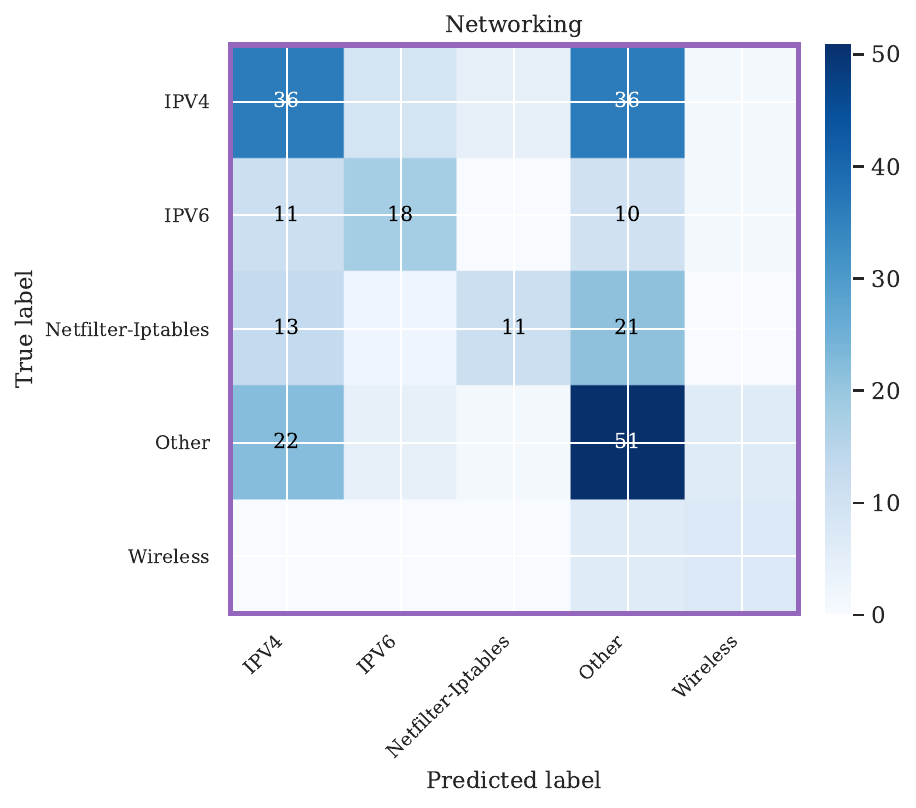}
        \caption{TabICL-Flat Networking Predictions}
    \end{subfigure}
    \caption{Subcategory confusion matrices for TabICL-Flat on the Bugs datset.}
    \label{fig:cmap_flat_sub}
\end{figure}

\begin{figure}[h]
    \centering
    \begin{subfigure}[t]{0.99\textwidth}
        \centering
        \includegraphics[width=\linewidth,height=\linewidth]{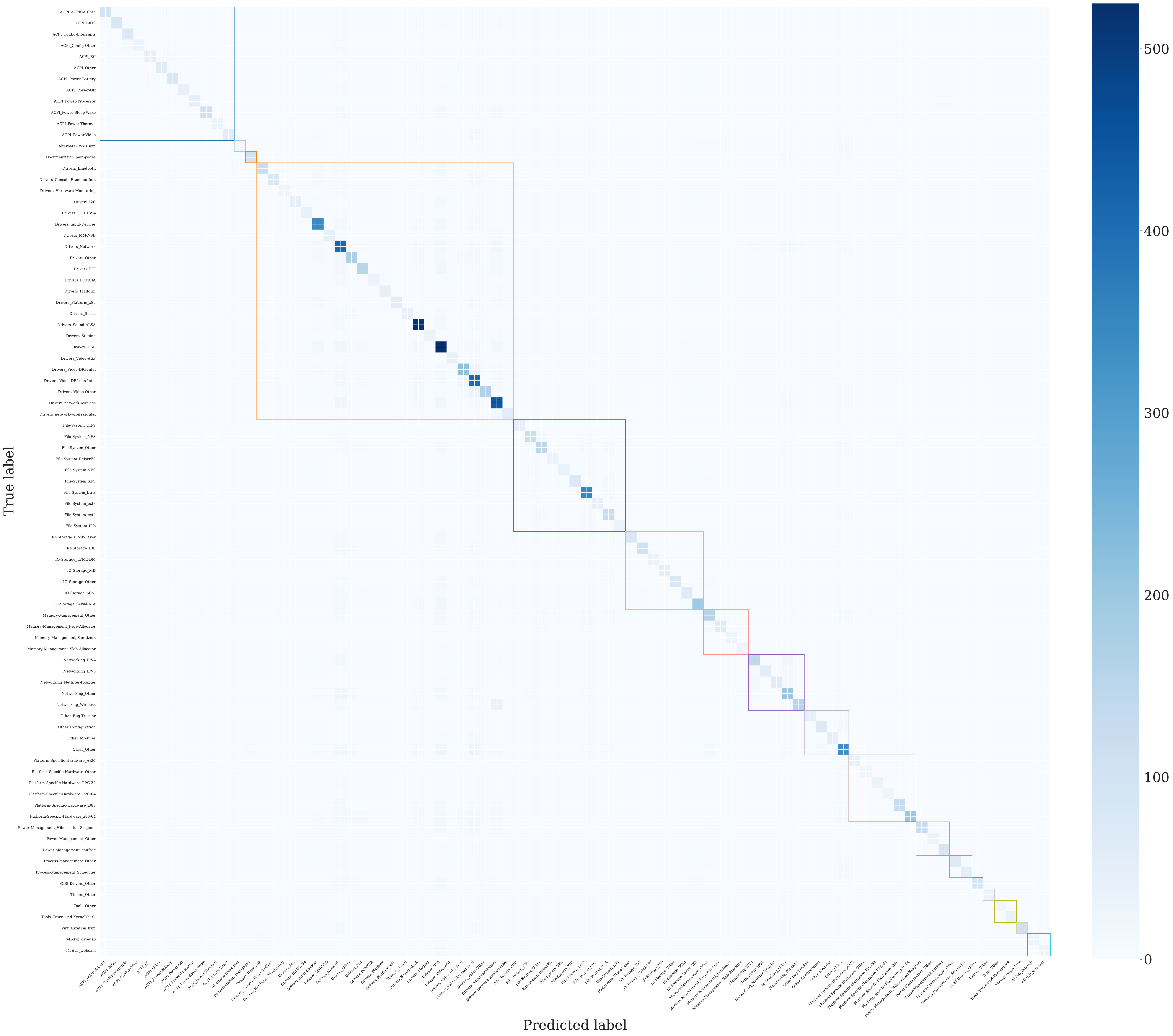}
        \caption{Original}
    \end{subfigure}

    \caption{Overview confusion matrix for TabICL-Hierarchical on the Bugs dataset. Coloured blocks represent subcategories, a selection of which are shown in Figure \ref{fig:cmap_h_sub}.}
    \label{fig:cmap_flat}
\end{figure}

\begin{figure}[h]
    \centering
    \begin{subfigure}[t]{0.49\textwidth}
        \centering
        \includegraphics[width=\linewidth,height=\linewidth]{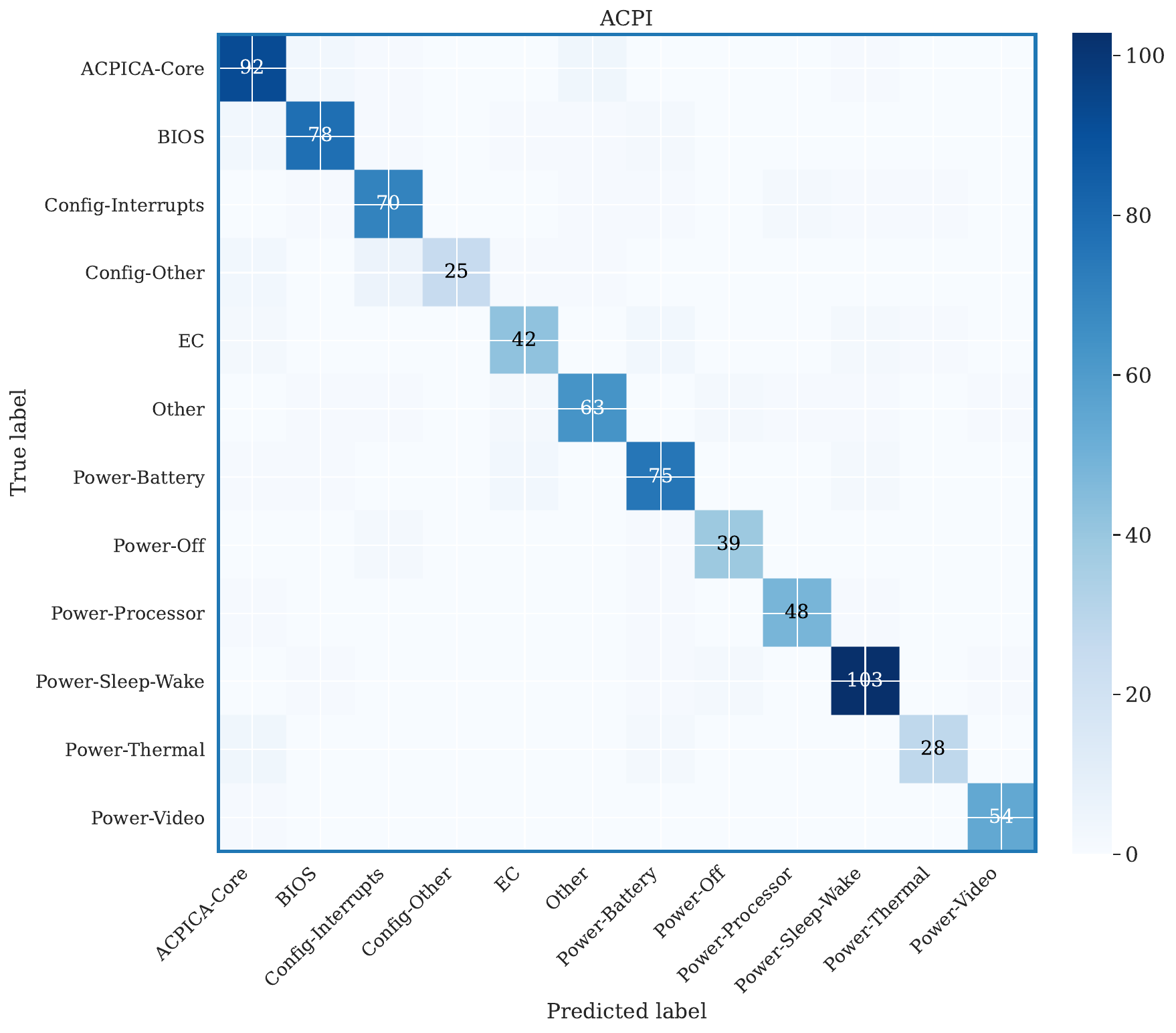}
        \caption{TabICL-Hierarchical ACPI Prediction}
    \end{subfigure}
    \hfill
    \begin{subfigure}[t]{0.49\textwidth}
        \centering
        \includegraphics[width=\linewidth,height=\linewidth]{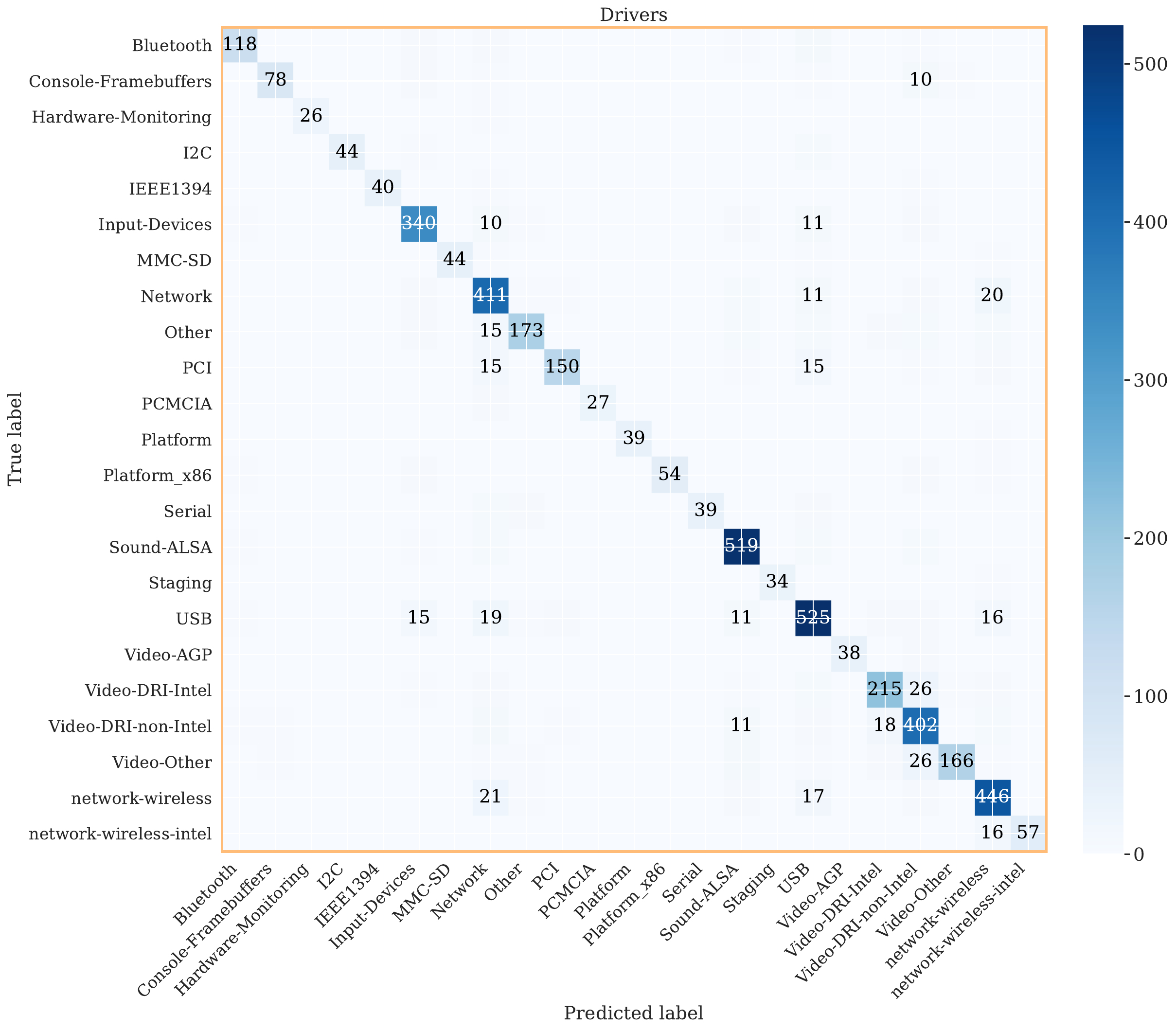}
        \caption{TabICL-Hierarchical Drivers Prediction}
    \end{subfigure}
    \hfill
    \begin{subfigure}[t]{0.49\textwidth}
        \centering
        \includegraphics[width=\linewidth,height=\linewidth]{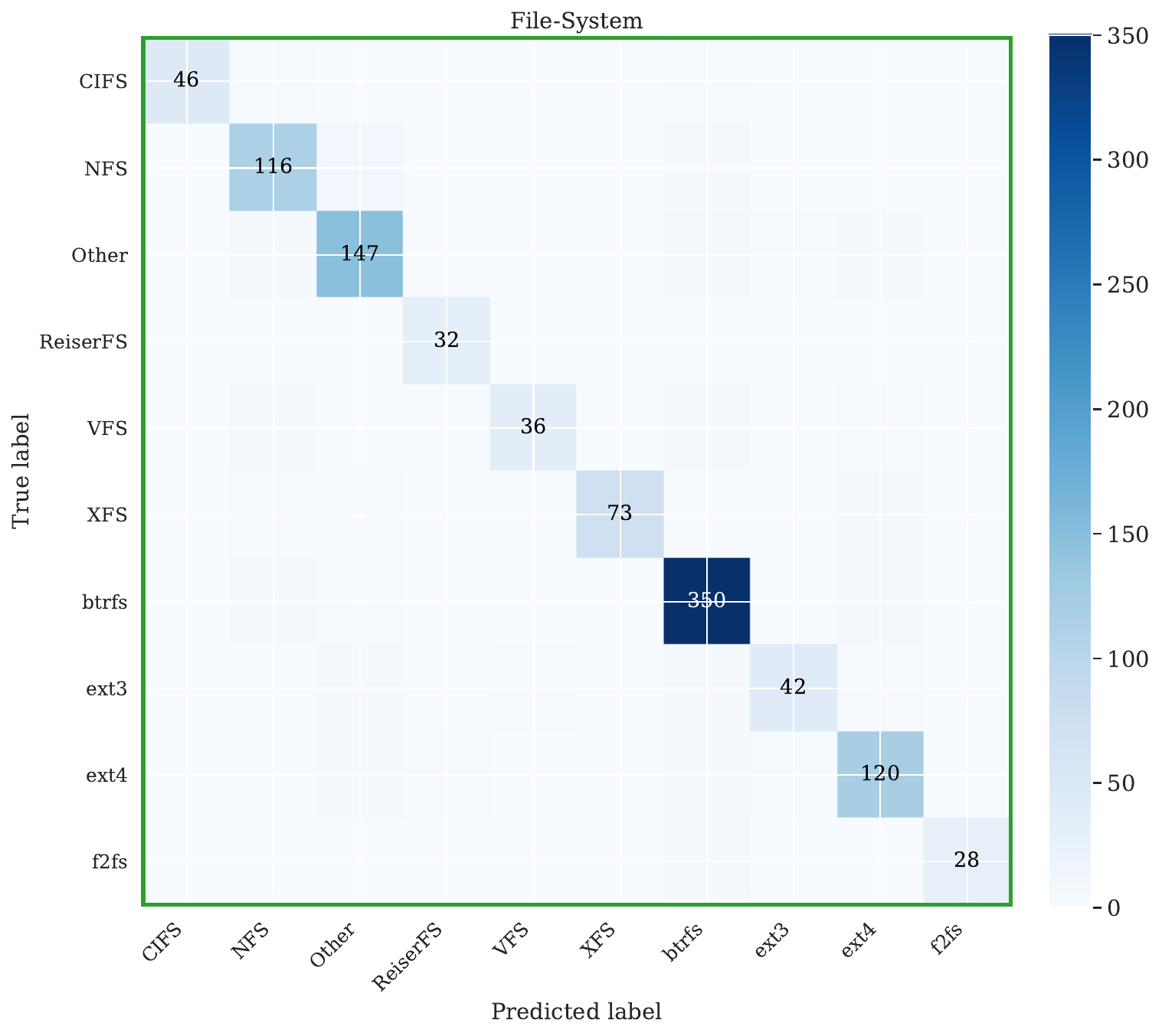}
        \caption{TabICL-Hierarchical File-System Predictions}
    \end{subfigure}
    \hfill
    \begin{subfigure}[t]{0.49\textwidth}
        \centering
        \includegraphics[width=\linewidth,height=\linewidth]{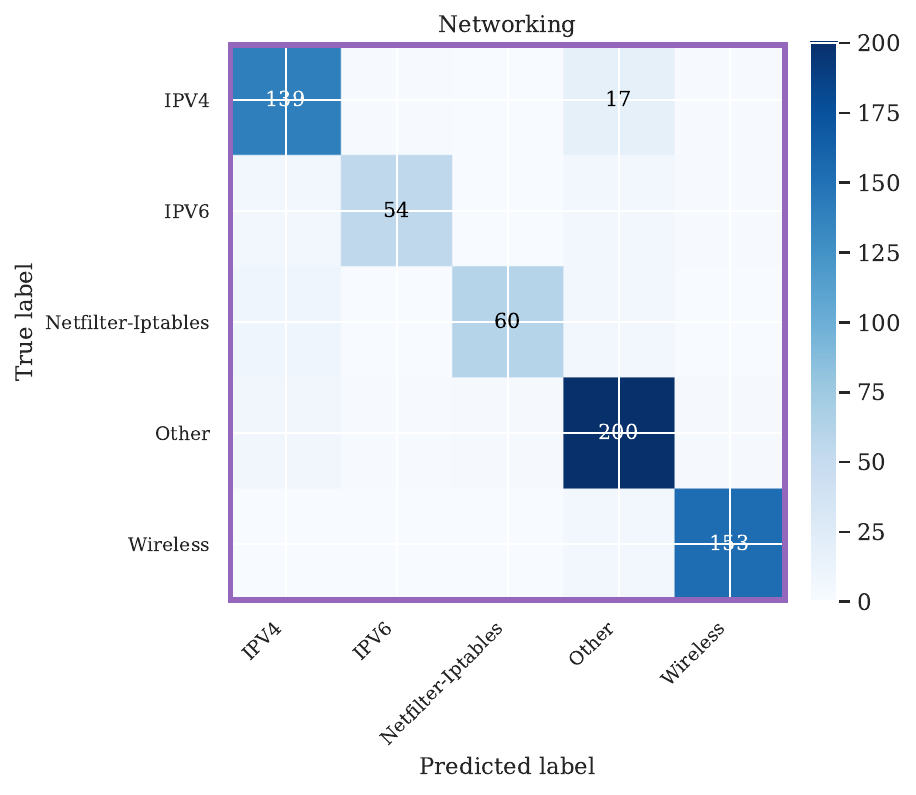}
        \caption{TabICL-Hierarchical Networking Predictions}
    \end{subfigure}
    \caption{Subcategory confusion matrices for TabICL-Hierarchical on the Bugs datset.}
    \label{fig:cmap_h_sub}
\end{figure}

\end{document}